\documentclass{eptcs}
\usepackage{mdframed}
\usepackage{microtype}
\usepackage[all=normal,floats=tight,mathspacing=tight]{savetrees}
\usepackage[T1]{fontenc}
\usepackage{color}
\usepackage{cite}
\usepackage{amsmath}
\usepackage{amssymb}
\usepackage{amsfonts}
\usepackage{algorithmic}
\usepackage{textcomp}
\usepackage{xcolor}
\usepackage[export]{adjustbox}
\usepackage{subfigure}
\usepackage{graphicx}
\usepackage{booktabs} % this is for making tables.
\usepackage{commath}
\usepackage{mathtools}
\usepackage{listings}
\usepackage{ifthen}
\usepackage{xurl}
\usepackage{float}
\usepackage{xstring}
\usepackage{tikz}
\usepackage{tikz-qtree}
\usepackage{xfp}
\usepackage{syntax}
\usepackage{caption}
\usepackage{bookmark}
\usepackage{newfloat} %needed for floating environment; grammar

\newboolean{IncludeAppendix}
\setboolean{IncludeAppendix}{true}
\newcommand{\CurrentChapter}{Null}
\newcommand{\CurrentSection}{Null}
\newcommand{\CurrentSubsection}{Null}
\newcommand{\CurrentSubsubsection}{Null}
\newcommand{\CurrentParagraph}{Null}

\newcommand{\LabelSection}[2]{#1::#2}
\newcommand{\LabelSubsection}[3]{#1::#2::#3}

\newcommand{\LabelParagraph}[5]{#1::#2::#3::#4::#5}
\newcommand{\LabelObject}[6]{#1::#2::#3::#4::#5::#6}

\newcommand{\LabelSubsectionHere}[1]{\LabelSubsection{\CurrentChapter}{\CurrentSection}{#1}}

\newcommand{\LabelObjectHere}[1]{\LabelObject{\CurrentChapter}{\CurrentSection}{\CurrentSubsection}{\CurrentSubsubsection}{\CurrentParagraph}{#1}}
\newcommand{\NewSection}[1]{\renewcommand{\CurrentSection}{#1}\renewcommand{\CurrentSubsection}{Null}\renewcommand{\CurrentSubsubsection}{Null}\renewcommand{\CurrentParagraph}{Null}\section{#1}\label{\LabelSection{\CurrentChapter}{#1}}\pdfbookmark[1]{#1}{\LabelSection{\CurrentChapter}{#1}}}
\newcommand{\NewSubsection}[1]{\renewcommand{\CurrentSubsection}{#1}\renewcommand{\CurrentSubsubsection}{Null}\renewcommand{\CurrentParagraph}{Null}\subsection{#1}\label{\LabelSubsection{\CurrentChapter}{\CurrentSection}{#1}}\pdfbookmark[2]{#1}{\LabelSubsection{\CurrentChapter}{\CurrentSection}{#1}}}

\newcommand{\NewParagraph}[1]{\renewcommand{\CurrentParagraph}{#1}\paragraph{#1}\label{\LabelParagraph{\CurrentChapter}{\CurrentSection}{\CurrentSubsection}{\CurrentSubsubsection}{#1}}\pdfbookmark[4]{#1}{\LabelParagraph{\CurrentChapter}{\CurrentSection}{\CurrentSubsection}{\CurrentSubsubsection}{#1}}}
\newcommand{\Integers}{\mathbb{Z}}
\newcommand{\SuchThat}{\text{ s.t. }}

\newcommand{\DefinedAs}{\triangleq}
\newcommand{\True}{\top}

\newcommand{\EmptySet}{\varnothing}

\newcommand{\Increment}{\mathbin{{+}{+}}}
\newcommand{\Decrement}{\mathbin{{-}{-}}}

\newcommand{\BehaviorTree}{\mathit{BT}}
\newcommand{\Invalid}{\mathit{I}}
\newcommand{\Failure}{\mathit{F}}
\newcommand{\Running}{\mathit{R}}
\newcommand{\Success}{\mathit{S}}

\definecolor{BehaviorTreeBlackboardColor}{HTML}{0000FF} %blue
\definecolor{BehaviorTreeEnvironmentColor}{HTML}{FF8C00} %dark orange
\newcommand{\EnvVar}[1]{\textcolor{BehaviorTreeEnvironmentColor}{#1}}
\newcommand{\BlVar}[1]{\textcolor{BehaviorTreeBlackboardColor}{#1}}

\usetikzlibrary{shapes.geometric, shapes.misc, shapes.arrows, shapes.callouts, shapes.multipart, fit, positioning, calc, matrix, automata}

\definecolor{BehaviorTreeSelectorColor}{HTML}{00FFFF}
\definecolor{BehaviorTreeSequenceColor}{HTML}{FFA500}
\definecolor{BehaviorTreeParallelColor}{HTML}{FFD700}
\definecolor{BehaviorTreeLeafColor}{HTML}{C0C0C0}

\tikzset{
    Selector/.style={
        draw=black,
        fill=BehaviorTreeSelectorColor,
        text=black,
        shape=chamfered rectangle,
        minimum size=15pt,
        inner sep=0pt,
        font=\tiny
    },
    Sequence/.style={
        draw=black,
        fill=BehaviorTreeSequenceColor,
        text=black,
        shape=rectangle,
        minimum size=15pt,
        inner sep=0pt,
        font=\tiny
    },
    Parallel/.style={
        draw=black,
        fill=BehaviorTreeParallelColor,
        text=black,
        shape=trapezium,
        trapezium left angle=60,
        trapezium right angle=120,
        minimum size=15pt,
        inner sep=0pt,
        trapezium stretches body,
        font=\tiny,
        align=center
    },
    Decorator/.style={
        draw=black,
        fill=white,
        text=black,
        shape=trapezium,
        trapezium left angle=67,
        trapezium right angle=67,
        minimum size=15pt,
        inner sep=0pt,
        trapezium stretches body,
        font=\tiny,
        align=center
    },
    Action/.style={
        draw=black,
        fill=BehaviorTreeLeafColor,
        text=black,
        shape=ellipse,
        minimum size=15pt,
        inner sep=0pt,
        font=\tiny
    },
    Check/.style={
        draw=black,
        fill=BehaviorTreeLeafColor,
        text=red,
        shape=ellipse,
        minimum size=15pt,
        inner sep=0pt,
        font=\tiny
    },
    Blackboard/.style={
        draw=blue,
        fill=white,
        text=black,
        shape=rectangle,
        font=\tiny,
        inner sep = .5pt
    },
  TreeTable/.style={
    matrix of nodes,
    row sep=0pt,
    column sep=0pt,
    nodes={
      rectangle,
      draw=black,
      align=center
    },
    minimum height=1pt,
    text depth=0.5ex,
    text height=1.5ex,
    nodes in empty cells,
    column 1/.style={
      nodes={text width=5.0em,font=\bfseries}
    }
  },
  TreeTableWide/.style={
    matrix of nodes,
    row sep=0pt,
    column sep=0pt,
    nodes={
      rectangle,
      draw=black,
      align=center
    },
    minimum width=7pt,
    minimum height=1pt,
    text depth=0.5ex,
    text height=1.5ex,
    nodes in empty cells,
    column 1/.style={
      nodes={text width=5.0em,font=\bfseries}
    }
  }
}
\newcommand{\CiteAsText}[1]{\cite{#1}}
\newcommand{\CiteAsRef}[1]{\cite{#1}}
\newcommand{\BiggestFishVar}{\BlVar{f}}

\newcommand{\FiniteStateMachine}{\mathit{FSM}}
\newcommand{\FSMState}{S_{FSM}}
\newcommand{\FSMInitial}{s_{FSM}}
\newcommand{\FSMAlphabet}{\Sigma_{FSM}}
\newcommand{\FSMTransition}{\delta_{FSM}}

\newcommand{\LTLFormula}{\phi}
\newcommand{\LTLFormulaA}{\phi_1}
\newcommand{\LTLFormulaB}{\phi_2}

\newcommand{\LTLNext}{\bigcirc}
\newcommand{\LTLUntil}{\mathcal{U}}
\newcommand{\LTLStrongRelease}{\mathcal{M}}
\newcommand{\LTLGlobally}{\square}
\newcommand{\LTLFinally}{\lozenge}

\newcommand{\Vertices}{V}
\newcommand{\Edges}{E}
\newcommand{\Root}{r}

\newcommand{\StatefulBehaviorTree}{\mathit{SBT}}
\newcommand{\SBTStates}{S_{SBT}}
\newcommand{\SBTInitialState}{s_{SBT}}
\newcommand{\SBTAlphabet}{\Sigma_{SBT}}
\newcommand{\SBTTransition}{\delta_{SBT}}
\newcommand{\StatusSet}{ST}
\newcommand{\Status}{st}
\newcommand{\VerticesPowerSeq}{VS}
\newcommand{\VerticesSeq}{vs}

\definecolor{DSLCommentColor}{HTML}{778899}
\newcommand{\DSLComment}[2]{\hspace{#1}\textcolor{DSLCommentColor}{\text{\##2}}}
\newcommand{\SBTFSM}{BTSM}

\newcommand{\SimpleRobotX}{\BlVar{x}}
\newcommand{\SimpleRobotY}{\BlVar{y}}
\newcommand{\SimpleRobotGoalX}{\EnvVar{x_g}}
\newcommand{\SimpleRobotGoalY}{\EnvVar{y_g}}
\newcommand{\SimpleRobotMGoalX}{\BlVar{x_g}}
\newcommand{\SimpleRobotMGoalY}{\BlVar{y_g}}

\DeclareFloatingEnvironment[
  fileext   = loc,
  listname  = List of Grammars,
  name      = Grammar,
  placement = htbp,
]{Grammar}

\definecolor{BehaVerifyLightColor1}{HTML}{D2691E} % Chocolate
\definecolor{BehaVerifyLightColor2}{HTML}{0000CD} % MediumBlue
\definecolor{BehaVerifyLightColor3}{HTML}{FF1493} % DeepPink
\definecolor{BehaVerifyLightColor4}{HTML}{FF1493} % DeepPink
\definecolor{BehaVerifyLightColor5}{HTML}{FF4500} % DarkOrange
\definecolor{BehaVerifyLightColor6}{HTML}{FF1493} % DeepPink
\definecolor{BehaVerifyLightColor7}{HTML}{FF1493} % DeepPink
\definecolor{BehaVerifyLightColor8}{HTML}{778899} % LightSlateGray

\definecolor{BehaVerifyLightColorBlack}{RGB}{0, 0, 0}
\definecolor{BehaVerifyLightColorWhite}{RGB}{255, 255, 255}

\lstdefinelanguage{BehaVerifyLight}{
    keywordstyle=[1]\color{BehaVerifyLightColor1},
    keywordstyle=[2]\color{BehaVerifyLightColor2},
    keywordstyle=[3]\color{BehaVerifyLightColor3},
    keywordstyle=[4]\color{BehaVerifyLightColor4},
    keywordstyle=[5]\color{BehaVerifyLightColor5},
    keywordstyle=[6]\color{BehaVerifyLightColor6},
    keywordstyle=[7]\color{BehaVerifyLightColor7},
    keywords=[1]{True, False, success, runnning, failure}, % constants such as true, false
    keywords=[2]{abs, max, min, sin, cos, exp, tan, ln, eq, neq, lt, gt, lte, gte, neg, add, sub, mult, idiv, rdiv, mod, count, index, not, and, or, xor, xnor, implies, equiv, active, success, running, failure, next, globally, globally_bounded, finally, finally_bounded, until, until_bounded, release, release_bounded, previous, not_previous_not, historically, historically_bounded, once, once_bounded, since, since_bounded, triggered, triggered_bounded, exists_globally, exists_next, exists_finally, exists_until, always_globally, always_next, always_finally, always_until}, % list all the functions here
    keywords=[3]{assign, end_assign, case, end_case, result, end_result, condition, end_condition, LTLSPEC, end_LTLSPEC, CTLSPEC, end_CTLSPEC, INVARSPEC, end_INVARSPEC},
    keywords=[4]{variable, end_variable, variable_statement, end_variable_statement, initial_values, end_initial_values, update, end_update, write_environment, end_write_environment, read_environment, end_read_environment, return_statement, end_return_statement},
    keywords=[5]{parallel, selector, sequence, X_is_Y, inverter, read_variables, end_read_variables, write_variables, end_write_variables, arguments, end_arguments, local_variables, end_local_variables, policy, success_on_all, success_on_one, with_partial_memory, with_true_memory, instant, constant_index, range, children, end_children, child, end_child, X, Y, bl, env, local, VAR, BOOLEAN},
    keywords=[6]{composite, end_composite, decorator, end_decorator, check, end_check, action, end_action, environment_check, end_environment_check, sub_tree, end_sub_tree, insert, end_insert}, % list the node categories here
    keywords=[7]{constants, end_constants, variables, end_variables, environment_update, end_environment_update, checks, end_checks, environment_checks, end_environment_checks, actions, end_actions, sub_trees, end_sub_trees, tree, end_tree, tick_prerequisite, end_tick_prerequisite, specifications, end_specifications}, % list the required sections here....hmm.....perhaps not? can't tell if it's a good idea. 
  sensitive=true,
  comment=[s]{\#\{}{\}\#},
  commentstyle=\color{BehaVerifyLightColor8}\slshape,
  stringstyle=\color{BehaVerifyLightColor1},
  morestring=[b]'
}

\lstdefinestyle{BehaVerifyLightStyle}{
  basicstyle=\tiny\color{BehaVerifyLightColorBlack}\ttfamily,
  backgroundcolor=\color{BehaVerifyLightColorWhite},
  language=BehaVerifyLight,
  numbers=left,
  numberstyle=\tiny\color{BehaVerifyLightColorBlack},
  stepnumber=1,
  numbersep=10pt
}\definecolor{nuXmvLightColor1}{RGB}{0,0,255} %blue
\definecolor{nuXmvLightColor2}{RGB}{255,51,51} %red
\definecolor{nuXmvLightColor3}{RGB}{0,204,102} %green
\definecolor{nuXmvLightColor4}{RGB}{153,153,0} %pale yellow
\definecolor{nuXmvLightColor5}{RGB}{255,153,204} %teal?
\definecolor{nuXmvLightColor6}{RGB}{0,102,51} %purple?
\definecolor{nuXmvLightColor7}{RGB}{0,200,200} %teal?
\definecolor{nuXmvLightColor8}{RGB}{255,255,0} %bright yellow (clashes with pale yellow).
\definecolor{nuXmvLightColor9}{RGB}{169,3,252}

\definecolor{nuXmvLightColorBlack}{RGB}{0, 0, 0}
\definecolor{nuXmvLightColorWhite}{RGB}{255, 255, 255}

\lstdefinelanguage{nuXmvLight}{
    keywordstyle=[1]\color{nuXmvLightColor1},
    keywordstyle=[2]\color{nuXmvLightColor2},
    keywordstyle=[3]\color{nuXmvLightColor3},
    keywordstyle=[4]\color{nuXmvLightColor4},
    keywordstyle=[5]\color{nuXmvLightColor5},
    keywordstyle=[6]\color{nuXmvLightColor6},
    keywordstyle=[7]\color{nuXmvLightColor7},
    keywords=[1]{TRUE, FALSE}, % constants such as true, false
    keywords=[2]{pi, abs, max, min, sin, cos, exp, tan, ln, xor, xnor, mod,  word1, bool, toint, count, swcost, uwconst, signed, unsigned, sizeof, floor, extend, resize, union, in, READ, WRITE, CONSTARRAY, typeof, EG, EX, EF, AG, AX, AF, U, X, G, F, V, Y, Z, H, O, S, T}, % list all the functions here
    keywords=[3]{ASSIGN, TRANS, INIT, FAIRNESS, JUSTICE, COMPASSION, INVAR, next, init, case, esac},
    keywords=[4]{CONSTANTS, DEFINE, VAR, IVAR, FROZENVAR, FUN},
    keywords=[5]{integer, real, boolean, array, unsigned word, signed word, clock},
    keywords=[6]{LTLSPEC, CTLSPEC, SPEC, INVARSPEC}, % list the node categories here
    keywords=[7]{MODULE}, % list the required sections here....hmm.....perhaps not? can't tell if it's a good idea. 
  sensitive=true,
  comment=[l]{--},
  commentstyle=\color{nuXmvLightColor9}
}

\lstdefinestyle{nuXmvLightStyle}{
  basicstyle=\tiny\color{nuXmvLightColorWhite}\ttfamily,
  backgroundcolor=\color{nuXmvLightColorBlack},
  language=nuXmvLight,
  numbers=left,
  numberstyle=\tiny\color{nuXmvLightColorBlack},
  stepnumber=1,
  numbersep=10pt,
  tabsize=2
}\definecolor{textXLightColor1}{HTML}{D2691E} % Chocolate
\definecolor{textXLightColor2}{HTML}{0000CD} % MediumBlue
\definecolor{textXLightColor3}{HTML}{FF1493} % DeepPink
\definecolor{textXLightColor4}{HTML}{FF1493} % DeepPink
\definecolor{textXLightColor5}{HTML}{FF4500} % DarkOrange
\definecolor{textXLightColor6}{HTML}{FF1493} % DeepPink
\definecolor{textXLightColor7}{HTML}{FF1493} % DeepPink
\definecolor{textXLightColor8}{HTML}{778899} % LightSlateGray

\definecolor{textXLightColorBlack}{RGB}{0, 0, 0}
\definecolor{textXLightColorWhite}{RGB}{255, 255, 255}

\lstdefinelanguage{textXLight}{
    keywordstyle=[1]\color{textXLightColor6},
    keywordstyle=[2]\color{textXLightColor8},
  keywords=[1]{STRICTFLOAT, FLOAT, INT, STRING, BOOL},
  sensitive=true,
  comment=[l]{//},
  commentstyle=\color{textXLightColor8}\slshape,
  stringstyle=\color{textXLightColor2},
  morestring=[b]'
}

\lstdefinestyle{textXLightStyle}{
  basicstyle=\tiny\color{textXLightColorBlack}\ttfamily,
  backgroundcolor=\color{textXLightColorWhite},
  language=textXLight,
  numbers=left,
  numberstyle=\tiny\color{textXLightColorBlack},
  stepnumber=1,
  numbersep=10pt,
  tabsize=2
}\lstdefinelanguage{boring}{
  morecomment=[l]{//},
  commentstyle=\color{gray}
}

\renewcommand{\CurrentChapter}{Formalizing Stateful Behavior Trees}
\renewcommand{\CurrentSection}{Null}
\renewcommand{\CurrentSubsection}{Null}
\renewcommand{\CurrentSubsubsection}{Null}
\renewcommand{\CurrentParagraph}{Null}

\begin{document}
\title{Formalizing Stateful Behavior Trees}

\author{Serena S. Serbinowska
  \institute{{0000{-}0002{-}9259{-}1586}\\Vanderbilt University\\ Nashville TN, USA}
  \email{serena.serbinowska@vanderbilt.edu}
  \and
  Preston Robinette
  \institute{{0000{-}0002{-}4906{-}2179}\\Vanderbilt University\\ Nashville TN, USA}
  \email{preston.robinette@vanderbilt.edu}
  \and
  Gabor Karsai
  \institute{{0000{-}0001{-}7775{-}9099}\\Vanderbilt University\\ Nashville TN, USA}
  \email{gabor.karsai@vanderbilt.edu}
  \and
  Taylor T. Johnson
  \institute{{0000{-}0001{-}8021{-}9923}\\Vanderbilt University\\ Nashville TN, USA}
  \email{taylor.johnson@vanderbilt.edu}
}

\def\titlerunning{Formalizing Stateful Behavior Trees} 
  
\def\authorrunning{S. Serbinowska, P. Robinette, G. Karsai, T. Johnson}

\maketitle
\vspace{-10pt}
\begin{abstract}
  Behavior Trees ($\BehaviorTree$s) are high-level controllers that are useful in a variety of planning tasks and are gaining traction in robotic mission planning. As they gain popularity in safety-critical domains, it is important to formalize their syntax and semantics, as well as verify properties for them. In this paper, we formalize a class of $\BehaviorTree$s we call Stateful Behavior Trees ($\StatefulBehaviorTree$s) that have auxiliary variables and operate in an environment that can change over time. $\StatefulBehaviorTree$s have access to persistent shared memory---often known as a blackboard---that keeps track of these auxiliary variables. We demonstrate that $\StatefulBehaviorTree$s are equivalent in computational power to Turing Machines when the blackboard can store mathematical (unbounded) integers. We also identify conditions where $\StatefulBehaviorTree$s have computational power equivalent to finite state automata, specifically where the auxiliary variables are of finitary types. We present a domain specific language (DSL) for writing $\StatefulBehaviorTree$s and adapt the tool BehaVerify for use with this DSL\@. This new DSL in BehaVerify supports interfacing with popular $\BehaviorTree$ libraries in Python, and also provides generation of Haskell code and nuXmv models, the latter of which are used for model checking temporal logic specifications for the $\StatefulBehaviorTree$s. We include examples and scalability results where BehaVerify outperforms another verification tool (MoVe4BT) by a factor of 100.

\end{abstract}

%
% abstract and title
%                                                                                                          1.0 pages
%\vspace{-20pt}
\NewSection{Introduction}
A Behavior Tree ($\BehaviorTree$) is a high-level controller that shares similarities with a hierarchical state machine, yet distinguishes itself by offering greater flexibility and modularity in defining behaviors\@. %it does? No idea what the limitations are on HSMs tbh ~Serena
At its core, a $\BehaviorTree$ organizes various behaviors within a tree structure, where leaf nodes encapsulate distinct behaviors and higher-level nodes define the control flow\@.
This hierarchical arrangement facilitates the design of complex behaviors and is both scalable and adaptable to changing circumstances or requirements\@.
\par
{$\BehaviorTree$}s were originally created for video game development and were devised to enhance the autonomy and realism of Non-Playable Characters (NPCs)\@.
An NPC is an entity within a video game that operates under the control of the game's Artificial Intelligence (AI)\@.
$\BehaviorTree$s are useful for specifying such behaviors\@.
The explainability and versatility of $\BehaviorTree$s have also led to their widespread adoption in areas like robotics and AI\@.
Accordingly, $\BehaviorTree$s have been used for a variety of tasks, such as for controlling wheeled-legged robots \CiteAsRef{de-luca2023IEEE-Robotics-and-Automation-Letters} and bipedal locomotion robots \CiteAsRef{gu2022ICRA}, in vision measurement systems of road users \CiteAsRef{qin2023IEEE-Transactions-on-Cybernetics}\@, and the management swarms \CiteAsRef{huang2022ICUS,jeong2022AIM}\@.
Additional applications can be found in this survey \CiteAsText{iovino2022Robotics-and-Autonomous-Systems}\@.
\par
As $\BehaviorTree$s continue to be adopted to address new and existing challenges in various domains, especially in real-world, safety-critical domains such as robotics, it is increasingly important to formalize their structure and behaviors\@.
Such formalization is crucial for the verification of safety and liveness specifications, ensuring the systems behave reliably and as intended under all conditions\@.
Toward this end, we provide a formalization we call Stateful Behavior Trees ($\StatefulBehaviorTree$s)\@.
$\StatefulBehaviorTree$s are a class of $\BehaviorTree$s that have auxiliary variables and operate in an environment\@.
The primary contributions of this work are the following.\@
\begin{enumerate}
\item
  We formalize a novel class of models we call $\StatefulBehaviorTree$s that operate in an environment and have global variables stored in persistent shared memory\@.
\item
  We demonstrate equivalence of $\StatefulBehaviorTree$s to Turing Machines and Finite State Automata under syntactic assumptions, which is of critical importance for model checking $\BehaviorTree$s\@.
\item
  We present a domain specific language (DSL) for writing $\StatefulBehaviorTree$s implemented in an entirely reworked software tool called BehaVerify \CiteAsRef{serena2022SEFM}\@.
\item
  We compare the entirely reworked BehaVerify \CiteAsRef{serena2022SEFM} to MoVe4BT \CiteAsRef{MoVe4BT} in different verification examples and outperform in each; in one, BehaVerify outperformed by a factor of over 100\@.
\end{enumerate}

%                                                                                                          1.0 pages
\NewSection{Related Work} % ---------------------------------------------------------------------- Section :: Related Work and Contributions
In this section, we discuss relevant literature, focusing on the verification of $\BehaviorTree$s and domain specific languages (DSLs) for $\BehaviorTree$s. We highlight the contributions of this paper within these contexts.
\NewParagraph{Our Prior Work}
In our prior work we presented BehaVerify~\CiteAsRef{serena2022SEFM}. That version of BehaVerify~\CiteAsRef{serena2022SEFM} took as input a Py~Trees~\CiteAsRef{PyTrees} object and walked the tree to create a nuXmv~\CiteAsRef{nuXmv} model\@.
The created nuXmv~\CiteAsRef{nuXmv} model was incomplete; it had composite and decorator nodes, but the leaf nodes were `stubs' for the user to fill in\@.
The same was true of variables\@.
The new version of BehaVerify~\CiteAsRef{serena2022SEFM} utilizes a Domain Specific Language (DSL)\@.
It takes as input a $\BehaviorTree$ specified using the DSL and produces as output a nuXmv~\CiteAsRef{nuXmv} model, a Py~Trees~\CiteAsRef{PyTrees} implementation of the $\BehaviorTree$, or a Haskell implementation of the $\BehaviorTree$\@.
Crucially, the nuXmv~\CiteAsRef{nuXmv} model is now complete; there are no `stubs' for the user to fill in as all the variables and leaf nodes are fully and completely generated\@.
\NewParagraph{Existing Behavior Tree Frameworks} % ---------------------------------------------------------------------- Paragraph :: Existing DSLs
To our knowledge, there are no existing DSLs for $\BehaviorTree$s, but there are several related libraries and frameworks\@.
\CiteAsText{ghzouli2023IEEE-Transactions-on-Software-Engineering} lists a variety of different DSLs, but we believe these would more correctly be classified as library implementations of $\BehaviorTree$s (e.g. BehaviorTree.CPP~\CiteAsRef{BehaviorTreeCPP} and Py~Trees~\CiteAsRef{PyTrees})\@.
% without a special-purpose language for describing $\BehaviorTree$s and how their state may be modified\@.
PROMISE~\CiteAsRef{garcia2020ICSE} is a DSL inspired by $\BehaviorTree$s, but it is not a DSL for $\BehaviorTree$s\@.
%vTSL (verifiable Task Specification Language) \CiteAsRef{heinzemann2018IROS} is a DSL for robotic tasks in general that could be extended for $\BehaviorTree$s, but was designed with a more traditional programming style in mind and is not, at present, a DSL for $\BehaviorTree$s\@.
MoVe4BT~\CiteAsRef{MoVe4BT}, which we compare against, uses an xml style for specifying $\BehaviorTree$s\@.
\NewParagraph{Verifying Behavior Trees}
There are several existing formal verification works for $\BehaviorTree$s\@.
\CiteAsText{biggar2020IEEE-Robotics-and-Automation-Letters} utilizes SPOT \CiteAsRef{lutz2022CAV} for verification of $\BehaviorTree$s, but is limited to atomic propositions and boolean operators\@.
Furthermore, the examples provided seemed to take over an hour to run for very small trees\@.
\CiteAsText{colledanchise2021IROS} does runtime verification for a fragment of Timed Propositional Temporal Logic (TPTL), but not design-time model verification\@.
We compared against BTCompiler in our previous paper \CiteAsText{serena2022SEFM}\@.                   
To the best of our knowledge, the only other existing and available tools for model verification of $\BehaviorTree$s are ArcadeBT~\CiteAsRef{henn2022FMICS} and MoVe4BT~\CiteAsRef{MoVe4BT}\@.
\par
ArcadeBT~\CiteAsRef{henn2022FMICS} is an automatic verification method for $\BehaviorTree$s that verifies safety properties by encoding the $\BehaviorTree$ using Linear Constrained Horn Clauses (LCHCs)\@.
To do this, ArcadeBT~\CiteAsRef{henn2022FMICS} includes an implementation of $\BehaviorTree$s in C++ that can be automatically converted to LCHCs and verified using Z3 \CiteAsRef{Z3}\@.
The tool has been evaluated on trees with up to 18 nodes\@.
By comparison, our new DSL and implementation in BehaVerify handles trees with 20000 nodes (see Section~\ref{\LabelSection{\CurrentChapter}{Verification Results for Stateful Behavior Trees}}) and supports verification of linear temporal logic (LTL) and computation tree logic (CTL) allowing for both liveness and safety to be verified\@.
\par
MoVe4BT~\CiteAsRef{MoVe4BT} allows for the verification of LTL specifications over nodes, but it cannot verify LTL specifications written as predicates over variables\@.
In contrast, the implementation of $\StatefulBehaviorTree$ verification in BehaVerify developed in this paper supports LTL specifications over variables and nodes\@.
MoVe4BT~\CiteAsRef{MoVe4BT} supports nodes with true parallelism, while BehaVerify does not\@.
However, Py~Trees~\CiteAsRef{PyTrees}, a popular implementation of $\BehaviorTree$s that BehaVerify targets, does not support true parallelism\@.
A more detailed comparison of the tools can be found in Section~\ref{\LabelSection{\CurrentChapter}{Verification Results for Stateful Behavior Trees}}, including experimental evaluations that demonstrate BehaVerify is able to verify trees 100 times bigger than MoVe4BT~\CiteAsRef{MoVe4BT} (20000 vs 200 nodes)\@.
\NewParagraph{BTs with State and Theoretical Foundation for Verification}
While not universal, it is common for $\BehaviorTree$s to interact with memory, referred to here as a \textit{blackboard}\@.
\CiteAsText{biggar2021IEEE-Robotics-and-Automation-Letters} compares pure $\BehaviorTree$s ($\BehaviorTree$s without a blackboard) to unrestricted $\BehaviorTree$s ($\BehaviorTree$s with a finite blackboard)\@.
While the unrestricted $\BehaviorTree$s are strictly more powerful than the pure $\BehaviorTree$s, \CiteAsRef{biggar2020arXiv} points out that this violates the `reactive' nature of $\BehaviorTree$s (\CiteAsText{biggar2020arXiv} states ``An architecture is reactive if its decision making depends only on the current state of the environment'')\@.
Instead of using blackboards, the authors of \CiteAsText{biggar2020arXiv} advocate for combining $\BehaviorTree$s with Stateful Components, thereby preserving the benefits of $\BehaviorTree$s without loss of computational power\@.
Regardless, practical major implementations of $\BehaviorTree$s (such as Py~Trees \CiteAsRef{PyTrees} and its Robotic Operating System (ROS) extension PyTreesRos, BehaviorTree.cpp \CiteAsRef{BehaviorTreeCPP}, and Unreal Engine \CiteAsRef{UnrealEngine}) all feature the blackboard\@.
As such, there is a practical need for a framework that addresses $\BehaviorTree$s with blackboards\@.
\par
Other researchers take a different approach and treat $\BehaviorTree$s as deterministic functions with control systems, as seen in the works by \CiteAsRef{ogren2022annurev}, \CiteAsRef{colledanchise2017IEEE}, \CiteAsRef{ogren2020IEEE-Robotics-and-Automation-Letters}, \CiteAsRef{sprague2022IEEE}, and \CiteAsRef{colledanchise2016IROS}, with \CiteAsText{marzinotto2014IEEE} analyzing the potential of $\BehaviorTree$s as alternatives to Controlled Hybrid Dynamical Systems\@.
While these undoubtedly describe $\BehaviorTree$s with state, there are crucial differences between this style and our formalization of $\StatefulBehaviorTree$s\@.
The issue at hand is that `state' is an ambiguous term; it could refer to either memory or to the environment\@.
For instance, it was not assumed that pure $\BehaviorTree$s do not function in a persistent environment; rather, the assumption was that the $\BehaviorTree$ does not leverage its own persistent memory to augment its behavior\@.
In essence, pure $\BehaviorTree$s are functional; if presented with the same set of inputs, they will produce the same outputs\@.
In this sense, the control system approach utilizes pure $\BehaviorTree$s without memory; the state represents the environment rather than memory\@.
In contrast, $\StatefulBehaviorTree$s consider both a blackboard \textit{and} the environment\@.
Moreover, the environment in the control system model was fully under control of the $\BehaviorTree$\@.
While this may be a reasonable assumption in certain contexts, it makes it impossible to model uncertainty (such as the presence of wind for drones)\@.
As such, our formalization allows for nondeterministic updates\@.
This is extended to the environment, which is allowed to change and develop according to user defined rules that can utilize nondeterminism\@.
\NewParagraph{Computational Power}
Finally, we consider the computational power of the resulting models\@.
\CiteAsText{colledanchise2014IEEE-RSJ} informally says $\BehaviorTree$s are the same as Finite State Machines ($\FiniteStateMachine$s), but does not explicitly state any assumptions, restrictions, etc\@.
On the other hand, \CiteAsText{biggar2021IEEE-Robotics-and-Automation-Letters} creates a hierarchy of Teleo-reactive programs (TR), Decision Trees (DT), $\BehaviorTree$s, and $\FiniteStateMachine$s\@.
They conclude that if you provide a TR, DT, or $\BehaviorTree$ with access to a finite blackboard that they can freely read from/write to, then they are equivalent to a $\FiniteStateMachine$\@.
Giving a $\FiniteStateMachine$ access to a finite blackboard does not increase the computational power of the model, as this is the equivalent of adding a finite number of additional states\@.
We take this a step further and consider the power of a $\StatefulBehaviorTree$ with an infinite blackboard (a blackboard capable of storing variables of unbounded size) and conclude that such a model has the computational power of a Turing Machine\@.

%                                                                                                          2.0 pages
\NewSection{Behavior Tree Overview}
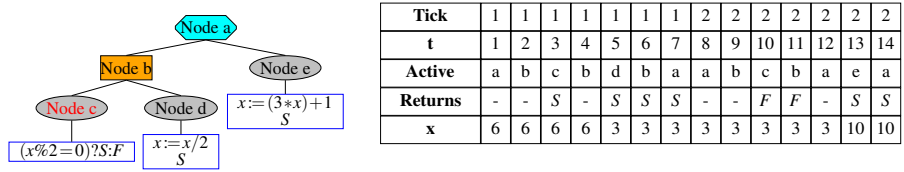
\begin{figure}%[b]
  %\vspace{-20pt}
  \centering
  \begin{adjustbox}{max width = 0.75\linewidth}
    \begin{tikzpicture}
  \tikzset{level distance=25pt}
  \tikzset{sibling distance=5pt}

  \Tree [.\node[Selector] (a) {\normalsize Node a};
  [.\node[Sequence] (b){\normalsize Node b};
  [.\node[Check] (c) {\normalsize Node c}; \node[Blackboard]{\begin{tabular}{c}\normalsize $(x \% 2 = 0)?\Success${:}$\Failure$\end{tabular}};]
  [.\node[Action] (d) {\normalsize Node d}; \node[Blackboard]{\begin{tabular}{c}\normalsize $x \coloneqq x / 2$\\\normalsize $\Success$\end{tabular}};]
  ]
  [.\node[Action] (e) {\normalsize Node e}; \node[Blackboard]{\begin{tabular}{c}\normalsize $x \coloneqq (3 * x) + 1$\\\normalsize $\Success$\end{tabular}};]
  ]

  \matrix (table) [TreeTable, text width={1em}, right={3.0cm of a}, yshift={-1cm}]
  {
    Tick      & 1 & 1 & 1 & 1 & 1 & 1 & 1 & 2 & 2 & 2 & 2 & 2 & 2 & 2\\
    t         & 1 & 2 & 3 & 4 & 5 & 6 & 7 & 8 & 9 & 10 & 11 & 12 & 13 & 14\\
    Active    & a & b & c & b & d & b & a & a & b & c & b & a & e & a\\
    Returns   & {-} & {-} & $\Success$ & {-} & $\Success$ & $\Success$ & $\Success$ & {-} & {-} & $\Failure$ & $\Failure$ & {-} & $\Success$ & $\Success$\\
    x         & 6 & 6 & 6 & 6 & 3 & 3 & 3 & 3 & 3 & 3 & 3 & 3 & 10 & 10\\
  };
\end{tikzpicture}
  \end{adjustbox}
  \caption{
    A $\BehaviorTree$ for the Collatz conjecture (hailstone sequence) consisting of a selector node (a), a sequence node (b), a check (c), and two actions (d, e)\@.
    We use the ternary operator $i?j:k$ to mean if $i$ then $j$ else $k$\@.
    We use \% for modulo\@.
    Tick indicates the number of times the tree has been ticked\@.
    $t$ is used to track the number of times we have changed active nodes\@.
    Active is used to track where we are in the tree\@.
    Returns indicates what the active node returns; a {-} indicates the node did not return\@.
    If a node is finished, then it returns one of $\Success$, $\Failure$, or $\Running$\@.
  }\label{\LabelObjectHere{Execution Example}}
  %\vspace{-1.5em}
\end{figure}
We will utilize Figure~\ref{\LabelObjectHere{Execution Example}} to provide an intuitive explanation of $\BehaviorTree$s\@.
We will then provide additional details\@.
\par{}
In Figure~\ref{\LabelObjectHere{Execution Example}} we have a basic $\BehaviorTree$ for calculating a hailstone sequence from a starting value\@.
To calculate such a sequence, start with a positive integer and use the following rules: if the number is even, divide by 2; otherwise, multiply by 3 and add 1\@.
The root of the tree (a) is a selector node\@.
As the name implies, this node `selects' a child\@.
In this case, we want to select either `divide by 2' or `multiply by 3 and add 1'\@.
(b) is a sequence node; aptly named once more, this node executes a sequence, aborting if a failure is encountered\@.
Here our sequence is `number is even' followed by `divide by 2'\@.
(c) is a check node; it checks if a condition is true or false\@.
(d) and (e) are action nodes; they actually do stuff\@.
\par{}
Our tree starts when it receives an external signal (a \textit{tick})\@.
This causes the root (a) to become active\@.
Execution will now, mostly, follow a \textit{depth first traversal} (DFT)\@.
(a) hasn't yet selected an option, so it follows DFT and makes (b) active\@.
(b) hasn't completed the sequence or encountered a failure, so it follows DFT and makes (c) active\@.
(c) checks if x is even; since 6 is even it returns Success ($\Success$)\@.
(b) is active again, but it still hasn't completed a sequence or encountered a failure, so it follows DFT and makes (d) active\@.
(d) executes the action and halves the value of x and returns $\Success$\@.
(b) is active again and the sequence successfully finished, so $\Success$ is returned\@.
(a) is finally active again and it has selected an option, so it returns $\Success$\@.
The first tick is now over\@.
The tree will now do nothing until it receives another tick\@.
When it does, (a) again becomes active, then (b), then (c)\@.\
This time, however, x is not even, so (c) returns Failure ($\Failure$)\@.
Thus the sequence failed, so (b) returns $\Failure$\@.
(a) is now active again, but still hasn't selected an option, so (e) becomes active\@.
(e) executes the action; x becomes 10 and $\Success$ is returned\@.
(a) has now selected an option, so it returns $\Success$\@.
The second tick is now over\@.
\par{}
We now provide some more concrete requirements for $\BehaviorTree$s that were not covered by the example\@.
There must be a path from the root to every other node in the tree\@.
Each node has exactly one parent, except the root which has no parent\@.
There are four possible states for nodes: Success ($\Success$), Failure ($\Failure$), Running ($\Running$), and Invalid ($\Invalid$)\@.
Each node becomes $\Invalid$ when a new tick arrives\@.
When a node finishes executing it returns one of $\Success$, $\Running$, or $\Failure$\@.
Finally, there are three types of nodes: leaf, decorator, and composite\@.
\NewParagraph{Leaf Nodes}
Leaf nodes are nodes that do not have any children, and they either check a condition (check node) or do an action (action node)\@.
Check nodes return $\Success$ if the associated condition is true and $\Failure$ otherwise; they do not do anything else\@.
Action nodes can perform various actions\@.
Furthermore, they are allowed to return $\Success$, $\Failure$, or $\Running$ and can utilize conditions to determine what to return\@.
In Figure~\ref{\LabelObject{\CurrentChapter}{\CurrentSection}{\CurrentSubsection}{\CurrentSubsubsection}{Null}{Execution Example}}, (c) is a check node while (d) and (e) are action nodes\@.
\NewParagraph{Decorator Nodes}
A decorator node always has exactly one child\@.
For our purposes, decorator nodes are used to change the output of a child node without requiring the child node to be modified\@.
Some common decorator types are inverter, which swaps $\Success$ and $\Failure$, and $\Running\_Is\_\Failure$ which turns $\Running$ into $\Failure$\@.
\NewParagraph{Composite Nodes}
Composite nodes control the execution flow through a $\BehaviorTree$\@.
There are three types of composite nodes: selector, sequence, and parallel\@.
The children are ordered and for convenience we will use a left-to-right order\@.
\begin{enumerate}
\item
  Selector or fallback nodes, try to `select' a child\@.
  A child is `selected' if it returns $\Success$\@.
  Each child is activated in order, from left to right, until one of them returns $\Success$ or $\Running$\@.
  At that point, the selector returns the same status\@.
  If every child returns $\Failure$, the selector returns $\Failure$\@.
\item
  Sequence nodes are identical to selector nodes, except $\Success$ and $\Failure$ are swapped\@.
  While this is true and useful to note, it is more practical to think of them in a more distinct manner\@.
  Sequence nodes are used to execute a sequence to the end or a failure point\@.
  Each child is activated in order, from left to right, until one of them returns $\Failure$ or $\Running$\@.
  At that point, the sequence returns the same status\@.
  If every child returns $\Success$, then the sequence returns $\Success$\@.
\item
  Parallel nodes will not appear in this paper, but it is still important to mention them\@.
  As the name implies, parallel nodes activate all their children simultaneously\@.
  We do not support this behavior and neither does Py~Trees \CiteAsRef{PyTrees}, the Python implementation that BehaVerify targets\@.
  Instead, our parallel nodes activate each child in order, one at a time, left to right\@.
  Unlike selector and sequence nodes, there is no early termination condition for parallel nodes; each child \textit{will} be activated\@.
  Once all the children have returned, the parallel node consults a policy to determine what to return\@.
\end{enumerate}

%                                                                                                          2.0 pages
\NewSection{Formal Definition of Stateful Behavior Trees}
Here we provide a formal definition of a $\StatefulBehaviorTree$\@.
In service of this task, we start by defining a tree\@.
A rooted tree is a triple $(\Vertices, \Root, \Edges)$ such that
\begin{itemize}
\item
  $\Vertices$ is a finite set representing the vertices of the tree\@.
\item
  $\Root \in \Vertices$ is a vertex representing the root\@.
\item
  Let $\VerticesPowerSeq$ be the set of all finite sequences $\VerticesSeq = [v_0, v_1, \ldots, v_n]$ such that $\forall j,k \in \Integers \SuchThat (0 \leq j, k \leq n), (v_j, v_k \in \Vertices \land (v_j = v_k \implies j = k))$\@.
  That is to say, the elements of the sequence are unique vertices\@.
  If $a$ and $b$ are elements in a sequence, we will use $<, >, \leq, \geq$ to indicate relative order of the sequence\@.
  For example $a < b$ means that $a$ appears before $b$ in the sequence\@.
\item
  $\Edges : \Vertices \mapsto \VerticesPowerSeq$ is a function from vertices to sequences of vertices (the children)\@.
  It must also meet the following requirements
  \begin{itemize}
  \item
    $\forall v \in \Vertices, \Root \notin \Edges(v)$ (the root has no parent)\@.
  \item
    $\forall v, v^{\prime} \in \Vertices, v \neq v^{\prime} \implies \Edges(v) \cap \Edges(v^{\prime}) = \EmptySet$\@.
    Each vertex has at most one parent\@.
  \item
    $\forall v \in \Vertices, \exists [v_0, v_1, \ldots, v_n] \SuchThat v_0 = \Root \land v_n = v \land \forall j \in \Integers \SuchThat 0 \leq j < n, v_{j+1} \in \Edges(v_j)$\@.\\
    There exists a path from the root to each vertex\@.
  \end{itemize}
  These conditions ensure that the tree is actually a tree\@.
\end{itemize}
\par
\NewSubsection{Stateful Behavior Tree}
A $\StatefulBehaviorTree$ is a tuple $(\Vertices, \Root, \Edges, \SBTStates, \SBTInitialState, \SBTAlphabet, \SBTTransition)$ such that
\begin{itemize}
\item
  $(\Vertices, \Root, \Edges)$ is a tree\@.
\item
  $\SBTStates$ is a set representing the possible states of the blackboard of $\StatefulBehaviorTree$\@.
  We discuss the implications of this set being infinite vs finite in Subsection~\ref{\LabelSubsectionHere{Translating Stateful Behavior Trees to Finite State Machines}}\@.
\item
  $\SBTInitialState \in \SBTStates$ is the initial state of the blackboard\@.
\item
  $\SBTAlphabet$ is a set representing the possible inputs (the environment)\@.
\item
  $\StatusSet$ is the set of all functions $\Status : \Vertices \mapsto \{\Success, \Running, \Failure, \Invalid\}$\@.
  Each $\Status \in \StatusSet$ is a function that maps each vertex to a status\@.
  $\StatusSet$ is not an element of the tuple; it arises from the elements\@.
\item
  $\SBTTransition : \Vertices \times \StatusSet \times \SBTStates \times \SBTAlphabet \mapsto 2^{\Vertices \times \StatusSet \times \SBTStates}$\@.
  Here $2^{\Vertices \times \StatusSet \times \SBTStates}$ is the power set of $\Vertices \times \StatusSet \times \SBTStates$\@.
  The function maps to sets to allow for the expression of nondeterminism\@.
  This function must also obey the following:
  \begin{equation*}
    \begin{split}
      \forall &v, v^{\prime} \in \Vertices, \forall  \Status, \Status^{\prime} \in \StatusSet, \forall s, s^{\prime} \in \SBTStates, \forall a \in \SBTAlphabet, (v^{\prime}, \Status^{\prime}, s^{\prime}) \in \SBTTransition(v, \Status, s, a) \implies\\
              &\begin{pmatrix*}
                (v = \Root \land \Status(v) \neq \Invalid \implies (v^{\prime} = \Root \land s = s^{\prime} \land (\forall v^{\prime\prime} \in \Vertices, \Status^{\prime}(v^{\prime\prime}) = \Invalid))) \land \\
                (v \neq \Root \implies \forall v^{\prime\prime} \in \Vertices, (v^{\prime\prime} = v) \lor \Status(v^{\prime\prime}) = \Status^{\prime}(v^{\prime\prime})) \land\\
                (v^{\prime} = v = \Root \lor v^{\prime} \in \Edges(v) \lor v \in \Edges(v^{\prime})) \land\\
                (v \in \Edges(v^{\prime}) \implies \Status^{\prime}(v) \neq \Invalid) \land\\
                (v^{\prime} \in \Edges(v) \implies \Status(v) = \Status^{\prime}(v) = \Status(v^{\prime}) = \Invalid) \land\\
                (v^{\prime} \in \Edges(v) \implies \forall v^{\prime\prime} \in \Edges(v), \Status(v^{\prime\prime}) \neq \Invalid \lor v^{\prime} \leq v^{\prime\prime})
              \end{pmatrix*}
    \end{split}
  \end{equation*}
  We will refer to $v$ as the active node and $v^{\prime}$ as the next node while explaining the above\@.
  If the root is active and not $\Invalid$, then we reset the status of the tree without changing anything else\@.
  In all other cases, only the status of the active node can be updated\@.
  The next node is either the root, the child of the active node, or the parent of the active node\@.
  If the next node is the parent of the active node, then the next status of the active node will not be $\Invalid$\@.
  If the active node is the parent of the next node, then the status of the active node and the next status of the active node are $\Invalid$ and the status of the next  node is $\Invalid$\@.
  If the active node is the parent of the next node, then all children that appear earlier in the sequence of active node's children are not $\Invalid$\@.
  These rules ensure that we move through the tree in the appropriate order\@.
\end{itemize}
Let $[a_0, a_1, \ldots]$ be a sequence of inputs from $\SBTAlphabet$\@.
Then a $\StatefulBehaviorTree$ trace is a sequence $[(v_0, \Status_0, s_0), (v_1, \Status_1, s_1), \ldots ]$ such that $v_0 = \Root$, $s_0 = \SBTInitialState$, $\forall v \in \Vertices, \Status_0(v) = \Invalid$, and $\forall j \in \Integers, j \geq 0 \implies (v_{j+1}, \Status_{j+1}, s_{j+1}) \in \SBTTransition(v_j, \Status_j, s_j, a_j)$\@.
\NewSubsection{Translating Stateful Behavior Trees to Finite State Machines}
Assuming that $\StatefulBehaviorTree$ has a finite alphabet set and a finite set of states, we will translate it into a nondeterministic Finite State Machine ($\FiniteStateMachine$)\@.
A $\FiniteStateMachine$ is a tuple $(\FSMState, \FSMInitial, \FSMAlphabet, \FSMTransition)$\@.
\begin{itemize}
\item
  $\FSMState$ is a set of states and $\FSMInitial \in \FSMState$ is the initial state\@.
\item
  $\FSMAlphabet$ is a set of possible inputs\@.
\item
  $\FSMTransition : \FSMState \times \FSMAlphabet \mapsto 2^{\FSMState}$ is the transition function\@.
  %Mapping to sets allows us to express nondeterminism\@.
\end{itemize}
Let $[a_0, a_1, \ldots]$ be a sequence of inputs from $\FSMAlphabet$\@.
Then a $\FiniteStateMachine$ trace is a sequence $[s_0, s_1, \ldots ]$ such that $s_0 = \FSMInitial$ and $\forall j \in \Integers, j \geq 0 \implies s_{j+1} \in \FSMTransition(s_j, a_j)$\@.
\par
Let $\StatefulBehaviorTree = (\Vertices, \Root, \Edges, \SBTStates, \SBTInitialState, \SBTAlphabet, \SBTTransition)$\@.
Assume $\SBTStates$ and $\SBTAlphabet$ are finite\@.
Furthermore, let $\StatusSet$ be the set of all functions $\Status : \Vertices \mapsto \{\Success, \Running, \Failure, \Invalid\}$ (as defined earlier)\@.
Since $\Vertices$ is a finite set, $\StatusSet$ is a finite set\@.
Then $\Vertices \times \SBTStates \times \StatusSet$ is finite as well\@.
Let $n$ be the number of elements in $\Vertices \times \SBTStates \times \StatusSet$\@.
Create a one-to-one mapping $\SBTFSM$ from $\Vertices \times \SBTStates \times \StatusSet$ to the integer interval $[0, 1, \ldots, n-1]$\@.
Let $\FSMTransition : [0, 1, \ldots, n-1] \times \SBTAlphabet \mapsto 2^{[0, 1, \ldots, n-1]}$ be defined such that
\begin{equation*}
  \begin{split}
    \forall v &\in \Vertices, \forall s \in \SBTStates, \forall \Status \in \StatusSet, \forall a \in \SBTAlphabet, \forall (v^{\prime}, s^{\prime}, \Status^{\prime}) \in \SBTTransition(v, s, \Status, a)\\
              &\SBTFSM(v^{\prime}, s^{\prime}, \Status^{\prime}) \in \FSMTransition(\SBTFSM(v, s, \Status), a).
  \end{split}
\end{equation*}
Then $([0, 1, \ldots, n-1], \SBTFSM(\Root, \SBTInitialState, \Status_0), \SBTAlphabet)$ is an equivalent $\FiniteStateMachine$, where $\Status_0 \in \StatusSet$ such that $\forall v \in \Vertices, \Status_0(v) = \Invalid$\@.
By translating the $\StatefulBehaviorTree$ to a $\FiniteStateMachine$, we allow for verification with tools such as nuXmv \CiteAsRef{nuXmv}\@.
\NewParagraph{Turing Complete}
In our translation, we assumed that $\SBTStates$ was finite\@.
If the blackboard can store one or more infinite variables (e.g.\ true integers), then $\SBTStates$ is not finite and $\StatefulBehaviorTree$s are Turing Complete\@.
To see this, consider the following:\@
\begin{itemize}
\item
  While the `tape' of a Turing Machine ($TM$) is infinite, the alphabet of symbols that can appear in each cell is finite\@.
  Assume that there are $n$ such symbols\@.
  Then the `tape' can clearly be represented as an integer in base $n$ where each cell is represented by a digit\@.
\item
  There are finitely many states that the Turing Machine can be in\@.
  Thus, these can be enumerated and stored using a finite integer\@.
\item
  The location of the `tape head' can be stored using an unbounded integer\@.
\end{itemize}
Thus by storing two unbounded integers and a bounded integer in the blackboard, we can fully capture the state and tape of a $TM$\@.
All that remains is to reproduce the transition function of the $TM$\@.
Because the transition function is a function from the set of finite states and finite alphabet to the set of finite states, finite alphabet, and a tape motion (Left, Right), it can easily be captured using a $\BehaviorTree$\@.
For each possible input to the transition function that has a defined output, create a 3-node subtree as seen in Figure~\ref{\LabelObjectHere{Turing Subtree}}\@.
Finally, add a selector root node with the 3-node subtrees as children\@.
\begin{figure}
  \begin{minipage}{.37\linewidth}
    \begin{adjustbox}{max width = \linewidth}
      \begin{tikzpicture}
  \tikzset{level distance=25pt}
  \tikzset{sibling distance=15pt}
  \Tree [.\node[Sequence](Seq){$Seq$};
    [.\node[Check](Chk){Check State}; \node[Blackboard]{\begin{tabular}{cc}$(sym = tape(h) \land$\\$st = st_{c})?\Success${:}$\Failure$\end{tabular}};]
    [.\node[Action](Act){Update State}; \node[Blackboard]{\begin{tabular}{cc}$tape(h) \coloneqq sym^{\prime}$\\$st_{c} \coloneqq st^{\prime}$\\$h \coloneqq h + dir$\\$\Success$\end{tabular}};]
  ]
  %\path (SymbolSequenceAZ) -- (SymbolSequenceZZ) node [black, font=\Large, midway, sloped] {$\dots$};
\end{tikzpicture}
    \end{adjustbox}
  \end{minipage}
  \begin{minipage}{.62\linewidth}
    \caption{
      Assume that $f(sym, st) = (sym^{\prime}, st^{\prime}, dir)$, where $f$ is the transition function for the Turing Machine\@.
      Then this means that if the Turing Machine is in state $st$ and reads $sym$ from the tape head, it will write $sym^{\prime}$ to the tape head, transition to $st^{\prime}$, and move the tape head according to $dir$\@.
      The subtree captures this behavior\@.
    }\label{\LabelObjectHere{Turing Subtree}}
  \end{minipage}
  %\vspace{-16pt}
\end{figure}
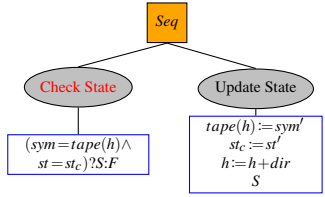
\NewParagraph{Turing Incomplete}
A $TM$ that only has access to a finite tape is not Turing Complete\@.
Similarly, if we restrict the blackboard to storing finitely many finite variables, the $\StatefulBehaviorTree$ ceases to be Turing Complete, as seen by the fact that a translation to a $\FiniteStateMachine$ exists\@.

%                                                                                                          3.0 pages
\NewSection{DSL and Implementation Details}
In this section, we provide details regarding the BehaVerify DSL used to specify $\StatefulBehaviorTree$s\@.
The grammar presented in Grammar~\ref{\LabelObjectHere{BehaVerify Grammar}} differs from the actual DSL (see \footnote{\url{https://github.com/verivital/behaverify/blob/main/metamodel/behaverify.tx}}) for the following reasons:
\begin{enumerate}
\item
  \textbf{Syntactic sugar.}
  For example, the actual DSL allows for fixed size arrays\@.
  In practice, this is equivalent to utilizing a large number of variables, but is more convenient (especially when combined with some basic loop functions)\@.
\item
  \textbf{Visual divisions.}
  The actual DSL uses far more pronounced visual dividers between sections (e.g., a case result is written as `case \{ Code \} result \{ Code, Code, \ldots \}')\@.
  This ensures that a specification written using the implementation is readable\@.
  However, it also injects a great deal of `text' into the grammar of the DSL, making it more difficult to parse here\@.
\item
  \textbf{Other features.}
  The actual DSL is still actively being developed for new features, many of which are not relevant to this paper (e.g., hyperproperties)\@.
  Such features were omitted\@.
\item
  \textbf{Format.}
  The actual DSL is written for use with textX~\CiteAsRef{textX} rather than with Backus-Naur Form\@.%; thus we do not define things like Int\@.
\end{enumerate}
\begin{mdframed}
  \begin{grammar}
  <SBT> ::= <Enums> `;' <Consts> `;' <BLVars> `;' <ENVVars> `;' <Env> `;' <Chks> `;' <Acts> `;' `{' <Node> `}' `;' <Specs> `;'

  % <ID> ::= <Letter><AlNum> \DSLComment{10pt}{Identifiers. Must start with a letter.}

  % <Letter> ::= `a' | `b' | `c' | \ldots | `z' | `A' | `B' | `C' | \ldots | `Z'

  % <Digit> ::= `0' | `1' | `2' | `3' | `4' | `5' | `6' | `7' | `8' | `9'

  % <AlNum> ::= <Letter> | <Digit> | <Letter><AlNum> | <Digit><AlNum>

  % <Int> ::= <Digit> | <Digit><Int> \DSLComment{10pt}{Actual DSL allows negatives.}\\\DSLComment{10pt}{For this example use (neg, Int) to get negatives} 

  % <Boolean> ::= `True' | `False'

  % <String> ::= `\textquotesingle' <AlNum> `\textquotesingle'

  <Status> ::= `success' | `running' | `failure'

  <Code> ::= <Int> | <Boolean> | <String> | <ID> | `(' <Function> `)' \\\DSLComment{10pt}{ID is used to reference variables, constants, etc}

  <CodeList> ::= <Code> | <Code>`,'<CodeList>

  <Function> ::= `add' `,' <Code> `,' <CodeList> | `not' `,' <Code> | \ldots

  <Enums> ::= $\varepsilon$ | `{' <String> `}' <Enums> \DSLComment{10pt}{$\varepsilon$ is the empty string}

  <Const> ::= <ID> `:=' <Int> | <ID> `:=' <Boolean> | <ID> `:=' <String>

  <Consts> ::= $\varepsilon$ | `{' <Const> `}' <Consts> \DSLComment{10pt}{$\varepsilon$ is the empty string}

  <Domain> ::= `BOOLEAN' | `[' <Code> `,' <Code> `]' | `{' <CodeList> `}' \\\DSLComment{10pt}{Code is used to allow constants and expressions}

  <CaseResult> ::= <Code> `?' <CodeList> \DSLComment{10pt}{If Case (left), then Result (right).}\\\DSLComment{10pt}{Choose nondeterministically if multiple Results}

  <Assign> ::= `{' <CaseResult>  `}' <Assign> | `{' <CodeList> `}' \\\DSLComment{10pt}{Try each CaseResult until one works. Default to CodeList if all fail}

  <RCaseResult> ::= <Code> `?' <Status> \\\DSLComment{10pt}{Same as CaseResult but for status. Deterministic.}

  <RAssign> ::= `{' <RCaseResult> `,' <RAssign> `}' | `{' <Status> `}'

  <Statement> ::= <ID> `,' <Assign> \DSLComment{10pt}{Update variable ID using Assign}

  <Statements> ::= $\varepsilon$ | `{' <Statement> `}' <Statements> \DSLComment{10pt}{$\varepsilon$ is the empty string}

  <Var> ::= <ID> `,' <Domain> `,' <Assign>

  <BLVars> ::= $\varepsilon$ | `{' <Var> `}' <BLVars> \DSLComment{10pt}{$\varepsilon$ is the empty string}

  <ENVVars> ::= <BLVars>

  <Envs> ::= <Statements>

  <Chk> ::= <ID> `,' <Code> \DSLComment{10pt}{Code must resolve to a Boolean}\\\DSLComment{10pt}{ID is the NodeType. Allows for reusage.}

  <Chks> ::= $\varepsilon$ | `{' <Chk> `}' <Chks> \DSLComment{10pt}{$\varepsilon$ is the empty string}

  <Act> ::= <ID> `,' <Statements> <RAssign> <Statements> \\\DSLComment{10pt}{ID is the NodeType. Allows for reusage.}

  <Acts> ::= $\varepsilon$ | `{' <Act> `}' <Acts> \DSLComment{10pt}{$\varepsilon$ is the empty string}

  <NodeBody> ::= `sequence' `{' <Children> `}' | `selector'  `{' <Children> `}'
  \alt `parallel one'  `{' <Children> `}' | `parallel all' `{' <Children> `}' 
  \alt `inverter' `{' <Node> `}' | `X\_is\_Y' <Status> <Status> `{' <Node> `}'
  \alt <ID> `:=' <ID> \DSLComment{5pt}{Name of leaf node (left) and NodeType (right)}

  <Node> ::= <ID> `,' <NodeBody>

  <Children> ::= `{' <Node> `}'  | `{' <Node> `}' <Children>

  <Spec> ::= `LTL' `{' <Code> `}' | `CTL' `{' <Code> `}' | `Invar' `{' <Code> `}'

  <Specs> ::= $\varepsilon$ | <Spec> <Specs> \DSLComment{10pt}{$\varepsilon$ is the empty string}
\end{grammar}
\captionsetup{hypcap=false}
\captionof{Grammar}{Representation of BehaVerify DSL with slight changes. We avoid defining basic types such as Int or ID (a letter followed by letters or digits).}\label{\LabelObjectHere{BehaVerify Grammar}}
\captionsetup{hypcap=true}
\end{mdframed}
%\vspace{1pt}
Note that both Grammar~\ref{\LabelObjectHere{BehaVerify Grammar}} and the actual DSL allow for nonsensical statements (e{.}g{.} (add, 1, \textquotesingle{}dog\textquotesingle{}))\@.
It is possible to create a grammar to exclude such cases, but the practical implementation proved both cumbersome to maintain and slow in practice\@.
Instead, we made a semantic checker for basic type checking along with other cases not covered by the grammar structure (e{.}g{.} ensuring that identifiers are unique)\@.
Listing~\ref{\LabelObjectHere{DSL Example}} is an example of how Grammar~\ref{\LabelObjectHere{BehaVerify Grammar}} would be used to create the $\StatefulBehaviorTree$ in Figure~\ref{\LabelObject{\CurrentChapter}{Fastforwarding Execution}{Null}{Null}{Specification Writing}{Stage Execution}}\@.
\begin{lstlisting}[frame=single, framesep=0pt, language=boring, caption={A basic example of a $\StatefulBehaviorTree$ specified using Grammar~\ref{\LabelObjectHere{BehaVerify Grammar}}}, label=\LabelObjectHere{DSL Example}]
  ;;//no enumerations or constants
  {x, [(neg, 1), 5], {0, 1}};
  //^Bl var x, nondeterministically initialized to 0 or 1
  {y, {0, 1}, {(eq, x, 0) ? 0} {1}};
  //^Env var y, initialized to 0 if x is 0, otherwise 1
  {y, {(eq, y, 0) ? 1} {0}};//between ticks, swap y value
  {cN, (geq, (add, x, y), 3)};
  //^declare check cN, checks if x+y is more than 3
  {aN1, {x, {(add, x, 1)}}, {success}}
  //^declare action aN1, adds 1 to x (; only on last action)
  {aN2, {x, {(sub, y, 1)}}, {success}};
  //^declare action aN2, set x to y-1
  {a, sequence {{b := aN1}{c := cN}{d := aN2}}};
  //^tree structure
  ;//no specifications
\end{lstlisting}
\vspace{-9pt}
\NewParagraph{Specifications}
The user may write specifications for the $\StatefulBehaviorTree$ using Linear Temporal Logic (LTL), Computational Tree Logic (CTL), or using Invariants over first order logic with standard connectives (and, or, etc.).
In the case of LTL and CTL, temporal functions may be used that are otherwise unavailable\@.
We provide a brief overview on LTL, as we use it in Section~\ref{\LabelSection{\CurrentChapter}{Verification Results for Stateful Behavior Trees}} to specify the desired outcomes and confirm that they occur (or provide a violation)\@.
\par{}
LTL operates on traces (sequences of states)\@.
Let $tr = [s_0, s_1, s_2, \ldots]$ be a trace for a $\FiniteStateMachine$\@.
When considering such a trace, time $t$ refers to $s_t$\@.
In general, we are interested in whether an LTL formula is true for the entire trace; this is the same as asking if the LTL formula is true at time 0\@.
Below we provide a minimal Grammar~\ref{\LabelObjectHere{LTL Grammar}} and then an overview of the presented functions and some syntactic sugar.
\begin{mdframed}
  \begin{grammar}
  <LTL> :: = <a> \DSLComment{10pt}{First Order Logic Formula}
  \alt \ensuremath{\neg} <LTL> | <LTL> \ensuremath{\lor} <LTL> \DSLComment{10pt}{Boolean operators; in practice we allow more operators}
  \alt \ensuremath{\LTLNext}(<LTL>) | (<LTL>)\ensuremath{\LTLUntil}(<LTL>) \DSLComment{10pt}{Temporal operators next and until}
\end{grammar}
\captionsetup{hypcap=false}
\captionof{Grammar}{Minimal LTL Grammar.}\label{\LabelObjectHere{LTL Grammar}}
\captionsetup{hypcap=true}
\end{mdframed}
\begin{itemize}
\item
  $\LTLNext(\LTLFormula)$ (next) is true at time $t$ if $\LTLFormula$ is true at time $t+1$\@.
\item
  $(\LTLFormulaA)\LTLUntil(\LTLFormulaB)$ (until) is true at time $t$ if $\exists t^{\prime\prime}$ such that $t \leq t^{\prime\prime}$ and $\LTLFormulaB$ is true at $t^{\prime\prime}$ and $\forall t^{\prime}$ such that $t \leq t^{\prime} < t^{\prime\prime}$, $\LTLFormulaA$ is true at $t^{\prime}$\@.
\item
  $(\LTLFormulaA)\LTLStrongRelease(\LTLFormulaB)$ (strong release) is true at time $t$ if $\exists t^{\prime\prime}$ such that $t \leq t^{\prime\prime}$ and $\LTLFormulaB$ is true at $t^{\prime\prime}$ and $\forall t^{\prime}$ such that $t \leq t^{\prime} \leq t^{\prime\prime}$, $\LTLFormulaA$ is true at $t^{\prime}$\@.
\item
  $\LTLGlobally(\LTLFormula)$ (globally) is true at time $t$ if $\forall t^{\prime}$ such that $t \leq t^{\prime}$, $\LTLFormula$ is true at time $t^{\prime}$\@.\
\item
  $\LTLFinally(\LTLFormula)$ (finally) is true at time $t$ if $\exists t^{\prime}$ such that $t \leq t^{\prime}$, $\LTLFormula$ is true at time $t^{\prime}$\@.
\end{itemize}

%                                                                                                          3.5 pages
\NewSection{Fastforwarding Execution}
Recall the example execution in Figure~\ref{\LabelObject{\CurrentChapter}{Behavior Tree Overview}{Null}{Null}{Null}{Execution Example}}\@.
The execution presented was intuitive and clear, but also highlighted a clear drawback: it took 14 time steps to complete 2 ticks in a 5 node tree\@.
This is not ideal for verification\@.
The total encoding, presented in our previous paper \CiteAsRef{serena2022SEFM}, was created to address this issue\@.
The experiments conducted in \CiteAsText{serena2022SEFM} clearly demonstrated the performance concerns associated with stepping through nodes one at a time and demonstrated that the total encoding is an effective method by which to mitigate this\@.
However, that encoding required the user to edit the resulting model by hand: an arduous task requiring not only expertise in nuXmv but in how BehaVerify encoded the model\@.
We have now addressed this issue (the user need only provide a specification file; everything else is handled automatically) and will explain our solution below\@.
\NewParagraph{Fastforwarding and a review of the total encoding}
\begin{figure}[b!]
  \fbox{
    \begin{minipage}{.97\linewidth}
      \begin{minipage}{.19\linewidth}
        \centering
    \begin{adjustbox}{max width = \linewidth}
      \begin{tikzpicture}
  \tikzset{level distance=25pt}
  \tikzset{sibling distance=10pt}
  \Tree [.\node[Sequence](a){\normalsize A};
    [.\node[Action](b){\normalsize B};]
    [.\node[Action](c){\normalsize C};]
  ]
\end{tikzpicture}
    \end{adjustbox}
    \caption{
      \\(L) 3 node tree\@.
      \\(R) total encoding\@.
      \\Note: $x?y:z$ means if $x$, then $y$, else $z$\@.
    }\label{\LabelObjectHere{3 Node Tree}}
  \end{minipage}
  \begin{minipage}{.8\linewidth}
    \begin{itemize}
    \item
      $act$ is a function that describes if a node is active or not during a given tick\@.
      $status$ is a function that describes the status of a node during a given tick\@.
    \item
      $act(A) \DefinedAs \True$, $act(B) \DefinedAs act(A)$, $act(C) \DefinedAs act(A) \land (status(B) = \Success)$
    \item
      $status(A) \DefinedAs
      \begin{cases}
        \Invalid \text{ if } \neg act(A)\\
        \Failure \text{ if } status(B) = \Failure \lor status(C) = \Failure\\
        \Running \text{ if } status(B) = \Running \lor status(C) = \Running\\
        \Success \text{ if } status(B) = \Success \land status(C) = \Success\\
      \end{cases}$
    \item
      Let $user_B$ be a function specified by the user that depends on variables and outputs a status\@.
      Then $status(B) \DefinedAs act(B)?user_B(vars):\Invalid$\@.
    \item
      Let $user_C$ be a function specified by the user that depends on variables and outpus a status\@.
      Then $status(C) \DefinedAs act(C)?user_C(vars):\Invalid$\@.
    \end{itemize}
  \end{minipage}
\end{minipage}
}
  \vspace{-1.75em}
  %\noindent\rule{\textwidth}{1pt}
\end{figure}
The goal is to handle the entire tick in one step, rather than stepping through each node one at a time\@.
To that end, the total encoding represented the status of each node as a function of other nodes\@.
This is shown in Figure~\ref{\LabelObjectHere{3 Node Tree}}\@.
Both the encoding and the process of automatically creating the total encoding were part of our prior work\@.
However, our prior work required the user to manually handle the creation of functions such as $user_B$ in nuXmv, along with creating appropriate variable updates in nuXmv\@.
These limitations were the result of:
\begin{enumerate}
\item
  A lack of a DSL\@.
  Our prior work handled existing Py Tree objects to create the tree\@.
  Unfortunately, this was not conducive to specifying how leaf nodes behave\@.
\item
  The complexity of variables in a total encoding\@.
  Suppose the variable $x$ is 0 at the start of the tick and 1 at the end; what value of $x$ does the function $user_C$ use?
\end{enumerate}
This is where our new work comes into play\@.
The user uses our DSL to specify leaf nodes, defining what status they return and how they change variables, and BehaVerify takes that information and automatically creates an improved total encoding allowing for the fastforwarding of execution, complete with variable updates and functions for leaf nodes, no additional input from the user required\@.
BehaVerify resolves the issue with variables through the use of variable stages\@.
Each variable has at least one stage\@.
For each possible update to a variable, an additional stage is created representing the variable after the update\@.
Thus, the number of stages a variable has is equal to the number of possible updates to that variable during a single tick plus one\@.
This can be seen in Figure~\ref{\LabelObject{\CurrentChapter}{\CurrentSection}{\CurrentSubsection}{\CurrentSubsubsection}{Specification Writing}{Stage Execution}}\@.
Each stage describes the value of a variable during a portion of the tick; BehaVerify tracks which stage the variable is in and references the appropriate stage in other functions, thus resolving the issue posed above\@.
\NewParagraph{Specification Writing}
We found that fastforwarding often simplifies specification writing\@.
Consider Figure~\ref{\LabelObjectHere{Stage Execution}}\@.
Suppose we want to write that during each tick, (d) returns $\Success$ (note that this specification is false; during the first tick (d) is $\Invalid$)\@.
With fastforwarding, this can be written as an invariant condition, namely $status(d) = \Success$\@.
If we are not using fastforwarding, we must write this using LTL
\vspace{-7pt}
\begin{equation*}
  \LTLGlobally(status(a) = \Invalid \implies ((status(a) = \Invalid) \LTLUntil (status(d) = \Success))).
  \vspace{-3pt}
\end{equation*}
This specification is far more complicated, because we now have to define the duration of a tick\@.
In this case, we accomplish this task by realizing that since (a) is the root, the tick ends when (a) returns\@.
As such, this specification says that it is always the case that if (a) is $\Invalid$, then (a) will stay $\Invalid$ until (d) returns $\Success$\@.
\par
Note that sequential properties can also be written with fastforwarding, though may require a little more forethought\@.
For instance, suppose we want to specify that (b) eventually returns $\Success$ and until that happens, (c) does not  returns $\Success$\@.
Since (b) occurs before (c) in the tree, this specification can be written as $status(c) \neq \Success \LTLUntil status(b) = \Success$\@.
The reverse, namely that (c) eventually returns $\Success$ and until that happens (b) does not return $\Success$\@, is slightly trickier\@.
One might be tempted to write $status(b) \neq \Success \LTLUntil status(c) = \Success$, but this accepts the case where (b) and (c) both return $\Success$ for the first time on the same tick\@.
To account for this, one should use $\LTLStrongRelease$ instead of $\LTLUntil$\@.
In general, this example illustrates the point well; it is entirely possible to describe sequential events when writing specifications for fastforwarded execution, but one must be mindful of the structure of the tree\@.
Finally, it is important to note that some sequential specifications are trivially true, and their verification is not a major objective for BehaVerify\@.
For instance, if (b) and (c) are both active during a tick, then because (b) is before (c) in a depth first traversal of the tree, node (b) was active before node (c)\@.
BehaVerify does generate a list of nodes in order of depth first traversal; if one wishes to confirm such specifications, they may consult this order\@.
\begin{figure}[t!]
  \fbox{
    \begin{minipage}{.97\linewidth} %all encompassing minipage for the whole thing
  \begin{minipage}{.68\linewidth}
    \begin{tabular}{|c|c|c|c|c|c|c|c|c|c|c|c|c|c|}
      \hline
      Tick         & 1 & 1 & 1 & 1 & 1 & 2 & 2 & 2 & 2 & 2 & 2 & 2 & 2\\
      \hline
      t            & 0 & 1 & 2 & 3 & 4 & 5 & 6 & 7 & 8 & 9 & 10 & 11 & 12 \\
      \hline
      Active       & a & b & a & c & a & a & b & a & c & a & d & a & a\\
      \hline
      Returns      & {-} & $\Success$ & {-} & $\Failure$ & $\Failure$ & {-} & $\Success$ & {-} & $\Success$ & {-} & $\Success$ & $\Success$ & {-}\\
      \hline
      $\BlVar{x}$  & 0 & 1 & 1 & 1 & 1 & 1 & 2 & 2 & 2 & 2 & 1 & 1 & 1\\
      \hline
      $\EnvVar{y}$ & 0 & 0 & 0 & 0 & 0 & 1 & 1 & 1 & 1 & 1 & 1 & 1 & 0\\
      \hline
    \end{tabular}
  \end{minipage}
  \begin{minipage}{.31\linewidth}
    \begin{adjustbox}{max width = \linewidth}
      \begin{tikzpicture}
  \tikzset{level distance = 25pt}
  \tikzset{sibling distance = 5pt}
  % Upper left node
  \node[draw, rectangle, align=center, anchor=north] at (-2, 1.75) {\normalsize \begin{tabular}{c}\textbf{Initial}\\$\BlVar{x_0} \in \{0, 1\}$\\$\EnvVar{y_0} \coloneqq \BlVar{x_0} = 0?0${:}$1$\end{tabular}};
  % Upper right node
  \node[draw, rectangle, align=center, anchor=north] at (2, 1.75) {\normalsize \begin{tabular}{c}\textbf{Between Ticks}\\$\EnvVar{y_1} \coloneqq \EnvVar{y_0} = 0?1${:}$0$\end{tabular}};
  \Tree [.\node[Sequence] (a) {\normalsize a}; 
  [.\node[Action] (b) {\normalsize b}; \node[Blackboard]{\normalsize \begin{tabular}{c}$\BlVar{x_1} \coloneqq \BlVar{x_0} + 1$\\$\Success$\end{tabular}};]
  [.\node[Check] (c) {\normalsize c}; \node[Blackboard]{\normalsize $\BlVar{x_1} + \EnvVar{y_0} \geq 3?\Success${:}$\Failure$};]
  [.\node[Action] (d) {\normalsize d}; \node[Blackboard]{\normalsize \begin{tabular}{c}$\BlVar{x_2} \coloneqq \EnvVar{y_0} - 1$\\$\Success$\end{tabular}};]
  ]
\end{tikzpicture}
    \end{adjustbox}
  \end{minipage}\\ % DO NOT REMOVE. It's ensuring that the next minipage is placed correctly
  \begin{minipage}{.14\linewidth}
    \begin{adjustbox}{max width = \linewidth}
      \begin{tabular}{|c|c|c|}
        \hline
        Tick         & 1 & 2\\
        \hline
        a & $\Failure$ & $\Success$\\
        \hline
        b &  $\Success$ & $\Success$\\
        \hline
        c & $\Failure$ & $\Success$\\
        \hline
        d & $\Invalid$ & $\Success$\\
        \hline
        $\BlVar{x_0}$ & 0 & 1\\
        \hline
        $\BlVar{x_1}$ & 1 & 2\\
        \hline
        $\BlVar{x_2}$ & 1 & 1\\
        \hline
        $\EnvVar{y_0}$ & 0 & 1\\
        \hline
        $\EnvVar{y_1}$ & 1 & 0\\
        \hline
      \end{tabular}
    \end{adjustbox}
  \end{minipage}
  \begin{minipage}{.85\linewidth}
    \caption{
      Here $\BlVar{x}$ is a blackboard variable and $\EnvVar{y}$ is an environment variable (an input)\@.
      The input changes between ticks, so $\EnvVar{y_1}$ does not appear in the tree\@.
      The upper table shows what stepping through the tree one node at a time would look like\@.
      The lower table shows what fastforwarding looks like\@.
      The tree itself includes stage subscripts that were added for readability;
      the user would normally create a tree without knowledge of the stages, as BehaVerify handles stage creation\@.
      Note that when fastforwarding, the value of $\BlVar{x_2}$ is set to that of $\BlVar{x_1}$ if (d) does not execute\@.
      Furthermore, note that the value of $\BlVar{x}$ at the end of the tick ($\BlVar{x_2}$) is the value of $\BlVar{x}$ at the start of the next tick ($\BlVar{x_0}$)\@.
    }\label{\LabelObjectHere{Stage Execution}}
  \end{minipage}
  \end{minipage}
  }
  \vspace{-2.5pt}
\end{figure}
\NewParagraph{Optimizations}
An immediate concern raised by introducing stages is how this effects model size\@.
After all, if we went from storing a single variable with 10 states to storing 3 variables with 10 states each, then the model is now spending 1000 states on this single variable\@.
Fortunately, through optimizations, we can generally avoid this issue\@.
Specifically, if a variable update is deterministic, then the stage is a function of the previous stage and doesn't increase the model size\@.
Furthermore, even if an update is nondeterministic, we can sometimes avoid an increase in model size\@.
Consider a very simple model with one variable, $x$, which is updated nondeterministically twice per tick\@.
Then, without optimizations, we would have three stages $x_0$, $x_1$, and $x_2$, and each stage would increase the size of the model\@.
Here $x_0$ is the value of $x$ at the start of the tick, $x_1$ the value after the first update, and $x_2$ the value after the second update\@.
Note that the next value of $x_0$ is the current value of $x_2$\@.
This is true for all variables; the next value of stage 0 is equal to the current value of the last stage\@.
Furthermore, each other stage depends only on current values\@.
In this example, since nothing depends on the value of $x_2$ other than the next value of $x_0$, we can safely remove the last stage and simply make it so the next value of $x_0$ is the value that $x_2$ would have been assigned\@.
It is important to note that if a user writes a specification that checks the value of $x$ at the end of a tick, this optimization \textit{would} change the result; we detect such cases and automatically disable the optimization for the variable\@.
Furthermore, if another variable or node depends on the value of $x_2$, this optimization would change the result and would therefore be disabled\@.
In such cases, we can instead try to combine the 0 stage with the 1 stage using a similar process, with similar caveats\@.
We can attempt this with the first and last stage of each variable\@.
% The four versions we consider are \NoOpt{} (do nothing), \LastOpt{} (try last stage removal), \FirstOpt{} (try first stage removal), and \FullOpt{} (try first stage removal then last stage removal)\@.
% These optimizations do not change the semantics of the model, but enable scalable verification\@.

%                                                                                                          2.0 pages
\NewSection{Verification Results for Stateful Behavior Trees}
\begin{figure}%[t!]
  \fbox{
  \begin{minipage}{.97\linewidth}
  \begin{minipage}{.45\linewidth}
    \begin{adjustbox}{max width = \linewidth}
      % requires mathtools, amsfonts
% requires Command -> Behavior_Tree_Draw
% requires Command -> Biggest Fish

\begin{tikzpicture}
  \tikzset{level distance=20pt}
  \tikzset{sibling distance=1pt}
  \Tree [.\node[Sequence](BigSeq){FishSeq};
    [.\node[Selector](se5){SelectFish};
      [.\node[Check](chk0){$f = 0$?}; \node[Blackboard]{$\BiggestFishVar = 0?\Success${:}$\Failure$};]
      [.\node[Check](chk1){$f = 1$?}; \node[Blackboard]{$\BiggestFishVar = 1?\Success${:}$\Failure$};]
      [.\node[Check](chk2){$f = 2$?}; \node[Blackboard]{$\BiggestFishVar = 2?\Success${:}$\Failure$};]
      [.\node[Check](chk3){$f = 3$?}; \node[Blackboard]{$\BiggestFishVar = 3?\Success${:}$\Failure$};]
      [.\node[Check](chk4){$f = 4$?}; \node[Blackboard]{$\BiggestFishVar = 4?\Success${:}$\Failure$};]
    ]
    [.\node[Action](Bigger){Bigger}; \node[Blackboard]{\begin{tabular}{l}$\BiggestFishVar \coloneqq min(999,$\\$\qquad\quad\BiggestFishVar + 1)$\\$\Success$\end{tabular}};]
  ]
\end{tikzpicture}
    \end{adjustbox}
  \end{minipage}
  \begin{minipage}{.52\linewidth}
    \begin{adjustbox}{max width = \linewidth}
      % requires mathtools, amsfonts
% requires Command -> Behavior_Tree_Draw
% requires Command -> Biggest Fish

\begin{tikzpicture}
  \tikzset{level distance=20pt}
  \tikzset{sibling distance=1pt}
  \Tree [.\node[Selector](move4btSel){MoVeSel};
  [.\node[Sequence](BigSeq){FishSeq};
  [.\node[Selector](se5){SelectFish};
      [.\node[Check](chk0){$f = 0$?}; \node[Blackboard]{$\BiggestFishVar = 0?\Success${:}$\Failure$};]
      [.\node[Check](chk1){$f = 1$?}; \node[Blackboard]{$\BiggestFishVar = 1?\Success${:}$\Failure$};]
      [.\node[Check](chk2){$f = 2$?}; \node[Blackboard]{$\BiggestFishVar = 2?\Success${:}$\Failure$};]
      [.\node[Check](chk3){$f = 3$?}; \node[Blackboard]{$\BiggestFishVar = 3?\Success${:}$\Failure$};]
      [.\node[Check](chk4){$f = 4$?}; \node[Blackboard]{$\BiggestFishVar = 4?\Success${:}$\Failure$};]
    ]
    [.\node[Action](Bigger){Bigger}; \node[Blackboard]{\begin{tabular}{l}$\BiggestFishVar \coloneqq min(999,$\\$\qquad\quad\BiggestFishVar + 1)$\\$\Success$\end{tabular}};]
  ]
  [.\node[Check](chk5){$MoVe4BTCheck$}; \node[Blackboard]{$\BiggestFishVar = 5?\Success${:}$\Failure$};]
  ]
\end{tikzpicture}
    \end{adjustbox}
  \end{minipage}
  \caption{
    Left is original, right is reworked for MoVe4BT\@.
    Number of nodes is counted using right tree (10 in this case)\@.
    $\BiggestFishVar$ is the size of the `biggest fish' so far\@.
    Each check returns $\Success$ if the $\BiggestFishVar$ is equal to its number and $\Failure$ otherwise\@.
    If any check returns $\Success$, the left half of the tree returns $\Success$ and $\BlVar{f}$ is incremented\@.
  }\label{\LabelObjectHere{Bigger Fish Trees}}
\end{minipage}
}
  %\vspace{-10pt}
\end{figure}
Here we present formal verification results and include comparisons to MoVe4BT~\CiteAsRef{MoVe4BT} that demonstrate BehaVerify scales significantly better in the size of the tree and overall state space\@.
Additionally, we present an interesting example demonstrating that BehaVerify is capable of finding complex counterexamples\@.
All results were generated on a Dell Inc. OptiPlex 7040 with 64 GiB of Memory with an Intel i7--6700 CPU @ 3.40GHz with 8 cores\@.
The code used and instructions for reproducing the results are available~\footnote{\url{https://github.com/verivital/behaverify/tree/main/REPRODUCIBILITY/2024_FMAS_SBT}}\@.
\par
We note here that BehaVerify takes a specification file written using the DSL and produces a nuXmv model\@.
The timing results for BehaVerify are based \textit{solely} on the time it takes nuXmv to run on the generated model\@.
We do not include the time it takes for BehaVerify to generate the nuXmv model\@.
This is because we wanted to compare our encoding to that of the competing tool, MoVe4BT, and we felt this was best demonstrated through a comparison of the model checking aspect\@.
However, we also note that the compile times do not meaningfully change the outcome of the results; compiling the simple robot experiment takes fractions of a second even with a 30 by 30 grid and the same is true for the bigger fish experiment with 1000 nodes\@.
\NewParagraph{Scaling Tree Experiment: Bigger Fish}
The bigger fish experiment (see Figure~\ref{\LabelObject{\CurrentChapter}{\CurrentSection}{\CurrentSubsection}{Null}{Null}{Bigger Fish Trees}}) scales the tree while the blackboard and environment are unchanged\@.
$\BiggestFishVar$ is an integer between 0 and 1000 inclusive and is initially 0\@.
The upper limit was increased for the tests on 10000 and 20000 nodes\@.
We intended to verify $\LTLFinally(\LTLGlobally(\BiggestFishVar = n))$ (variable attains and maintains a value), but MoVe4BT does not support LTL specifications over variables; you are restricted to checking the statuses of nodes\@.
We modified the tree and created a node (named $MoVe4BTCheck$) to check the value of the variable (see Figure~\ref{\LabelObject{\CurrentChapter}{\CurrentSection}{\CurrentSubsection}{Null}{Null}{Bigger Fish Trees}} for details)\@.
We then tried to verify $\LTLFinally(\LTLGlobally(MoVe4BTCheck = \Success))$\@.
In BehaVerify, this specification states that eventually, during each tick the check returns $\Success$\@.
However, because MoVe4BT does not utilize fastforwarding (see Section~\ref{\LabelSection{\CurrentChapter}{Fastforwarding Execution}} for details), MoVe4BT interprets this to mean that eventually the check is always active, which is false\@.
% Without fastforwarding, writing tick-oriented specifications is complicated\@.
We instead had to settle for verifying $\LTLFinally(MoVe4BTCheck = \Success)$\@.
This means that instead of verifying that the variable attains the value and maintains it, MoVe4BT only verifies that it attains it\@.
It is crucial to stress that BehaVerify is capable of verifying the original condition; in fact, we include timing results for both the original and new condition\@.
Figure~\ref{\LabelObjectHere{Biggest Fish Timing Results}} shows BehaVerify scales well with tree complexity, while MoVe4BT does not\@.
The specification is true\@.
If the model is changed so it is false (by removing a leaf in the chain), both tools produce a counterexample\@.
\begin{figure}%[t!]
  \begin{minipage}{.5\linewidth}
    \includegraphics[width=\linewidth,keepaspectratio]{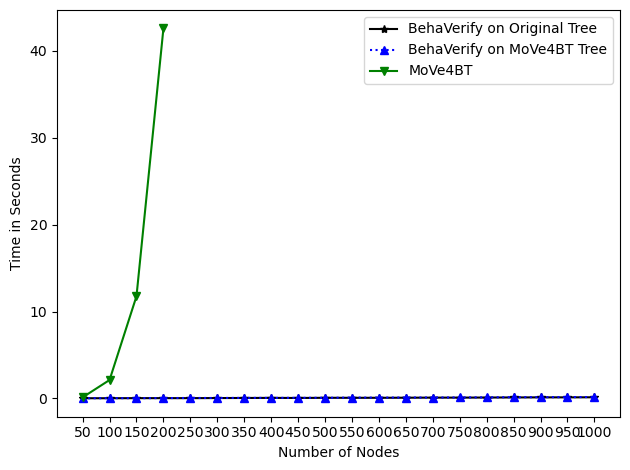}
  \end{minipage}
  \begin{minipage}{.49\linewidth}
    \caption{
      Time to verify $\LTLFinally(\LTLGlobally(\BiggestFishVar = n))$ (original) and $\LTLFinally(\BiggestFishVar = n)$ (changed) where $n$ is the number of nodes minus 5\@.
      Starting at 250 nodes total, MoVe4BT ran for over a minute, producing a blank screen with no results\@; we interpret this as a timeout\@.
      We ran BehaVerify with 10000 and 20000 nodes, taking 8.20 and 32.32 seconds\@.
      At 200 and 20000 nodes, BehaVerify reported 196 and 19996 reachable states\@.
      BehaVerify reports similar runtimes for both trees\@.
    }\label{\LabelObjectHere{Biggest Fish Timing Results}}
    \end{minipage}
    %\vspace{-1em}
\end{figure}
\NewParagraph{Scaling Blackboard Experiment: Simple Robot}
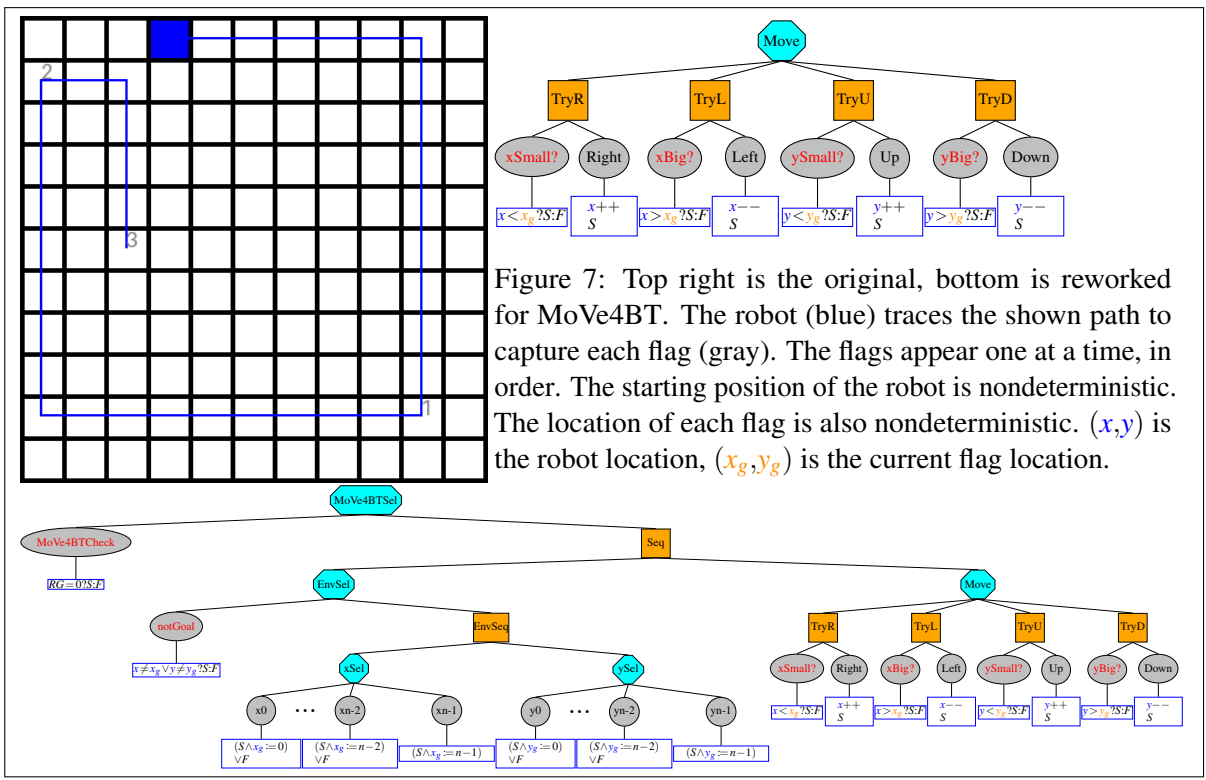
\begin{figure}%[t!]
\fbox{
\begin{minipage}{.97\linewidth}
  \begin{minipage}{.4\linewidth}
    \includegraphics[width=\linewidth,keepaspectratio]{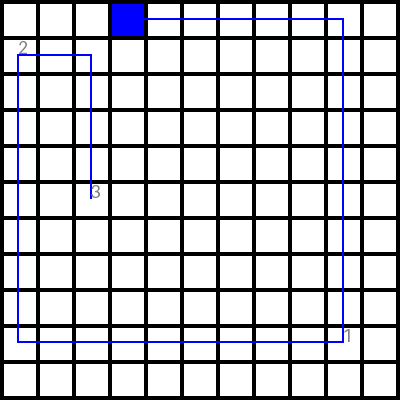}
  \end{minipage}
  \begin{minipage}{.58\linewidth}
    \begin{adjustbox}{max width = \linewidth}
      % requires mathtools, amsfonts
% requires Command -> Behavior_Tree_Draw
% requires Command -> Simple_Robot

\begin{tikzpicture}
  \tikzset{level distance=22pt}
  \tikzset{sibling distance=0.5pt}
  \Tree [.\node[Selector](move){Move};
    [.\node[Sequence](tryR){TryR};
      [.\node[Check](smallX){xSmall?}; \node[Blackboard]{$\SimpleRobotX < \SimpleRobotGoalX?\Success${:}$\Failure$};]
      [.\node[Action](goR){Right}; \node[Blackboard]{\begin{tabular}{l}$\SimpleRobotX \Increment$\\$\Success$\end{tabular}};]
    ]
    [.\node[Sequence](tryL){TryL};
      [.\node[Check](bigX){xBig?}; \node[Blackboard]{$\SimpleRobotX > \SimpleRobotGoalX?\Success${:}$\Failure$};]
      [.\node[Action](goL){Left}; \node[Blackboard]{\begin{tabular}{l}$\SimpleRobotX \Decrement$\\$\Success$\end{tabular}};]
    ]
    [.\node[Sequence](tryU){TryU};
      [.\node[Check](smallY){ySmall?}; \node[Blackboard]{$\SimpleRobotY < \SimpleRobotGoalY?\Success${:}$\Failure$};]
      [.\node[Action](goU){Up}; \node[Blackboard]{\begin{tabular}{l}$\SimpleRobotY \Increment$\\$\Success$\end{tabular}};]
    ]
    [.\node[Sequence](tryD){TryD};
      [.\node[Check](bigY){yBig?}; \node[Blackboard]{$\SimpleRobotY > \SimpleRobotGoalY?\Success${:}$\Failure$};]
      [.\node[Action](goD){Down}; \node[Blackboard]{\begin{tabular}{l}$\SimpleRobotY \Decrement$\\$\Success$\end{tabular}};]
    ]
  ]
\end{tikzpicture}
    \end{adjustbox}
    \caption{
      Top right is the original, bottom is reworked for MoVe4BT\@.
      The robot (blue) traces the shown path to capture each flag (gray)\@.
      The flags appear one at a time, in order\@.
      The starting position of the robot is nondeterministic\@.
      The location of each flag is also nondeterministic\@.
      $(\BlVar{x}, \BlVar{y})$ is the robot location, $(\EnvVar{x_g}, \EnvVar{y_g})$ is the current flag location\@. 
    }\label{\LabelObjectHere{Simple Robot Run and Trees}}
  \end{minipage}\\ %  DO NOT REMOVE, it's ensuring the next minipage is placed correctly.
  \begin{minipage}{\linewidth}
    \begin{adjustbox}{max width = \linewidth}
      % requires mathtools, amsfonts
% requires Command -> Behavior_Tree_Draw
% requires Command -> Simple_Robot

\begin{tikzpicture}
  \tikzset{level distance=22pt}
  \tikzset{sibling distance=0.5pt}
  \Tree [.\node[Selector](move4btsel){MoVe4BTSel};
  [.\node[Check](mcheck){MoVe4BTCheck}; \node[Blackboard]{$RG = 0?\Success${:}$\Failure$};]
  [.\node[Sequence](worldSeq){Seq};
  [.\node[Selector](envSel){EnvSel};
  [.\node[Check](notAtGoal){notGoal}; \node[Blackboard]{$\SimpleRobotX \neq \SimpleRobotMGoalX \lor \SimpleRobotY \neq \SimpleRobotMGoalY?\Success${:}$\Failure$};]
  [.\node[Sequence](envSeq){EnvSeq};
  [.\node[Selector](xSel){xSel};
  [.\node[Action](x0){x0};\node[Blackboard]{\begin{tabular}{l}$(\Success \land \SimpleRobotMGoalX \coloneqq 0)$\\$\lor \Failure$\end{tabular}};]
  [.\node[Action](xn2){xn-2};\node[Blackboard]{\begin{tabular}{l}$(\Success \land \SimpleRobotMGoalX \coloneqq n-2)$\\$\lor \Failure$\end{tabular}};]
  [.\node[Action](xn1){xn-1};\node[Blackboard]{\begin{tabular}{l}$(\Success \land \SimpleRobotMGoalX \coloneqq n-1)$\end{tabular}};]
  ]
  [.\node[Selector](ySel){ySel};
  [.\node[Action](y0){y0};\node[Blackboard]{\begin{tabular}{l}$(\Success \land \SimpleRobotMGoalY \coloneqq 0)$\\$\lor \Failure$\end{tabular}};]
  [.\node[Action](yn2){yn-2};\node[Blackboard]{\begin{tabular}{l}$(\Success \land \SimpleRobotMGoalY \coloneqq n-2)$\\$\lor \Failure$\end{tabular}};]
  [.\node[Action](yn1){yn-1};\node[Blackboard]{\begin{tabular}{l}$(\Success \land \SimpleRobotMGoalY \coloneqq n-1)$\end{tabular}};]
  ]
  ]
  ]
  [.\node[Selector](move){Move};
    [.\node[Sequence](tryR){TryR};
      [.\node[Check](smallX){xSmall?}; \node[Blackboard]{$\SimpleRobotX < \SimpleRobotGoalX?\Success${:}$\Failure$};]
      [.\node[Action](goR){Right}; \node[Blackboard]{\begin{tabular}{l}$\SimpleRobotX \Increment$\\$\Success$\end{tabular}};]
    ]
    [.\node[Sequence](tryL){TryL};
      [.\node[Check](bigX){xBig?}; \node[Blackboard]{$\SimpleRobotX > \SimpleRobotGoalX?\Success${:}$\Failure$};]
      [.\node[Action](goL){Left}; \node[Blackboard]{\begin{tabular}{l}$\SimpleRobotX \Decrement$\\$\Success$\end{tabular}};]
    ]
    [.\node[Sequence](tryU){TryU};
      [.\node[Check](smallY){ySmall?}; \node[Blackboard]{$\SimpleRobotY < \SimpleRobotGoalY?\Success${:}$\Failure$};]
      [.\node[Action](goU){Up}; \node[Blackboard]{\begin{tabular}{l}$\SimpleRobotY \Increment$\\$\Success$\end{tabular}};]
    ]
    [.\node[Sequence](tryD){TryD};
      [.\node[Check](bigY){yBig?}; \node[Blackboard]{$\SimpleRobotY > \SimpleRobotGoalY?\Success${:}$\Failure$};]
      [.\node[Action](goD){Down}; \node[Blackboard]{\begin{tabular}{l}$\SimpleRobotY \Decrement$\\$\Success$\end{tabular}};]
    ]
  ]
    ]
    ]
  \path (x0) -- (xn2) node [black, font=\Large, midway, sloped] {$\dots$};
  \path (y0) -- (yn2) node [black, font=\Large, midway, sloped] {$\dots$};
\end{tikzpicture}
    \end{adjustbox}
  \end{minipage}
  \end{minipage}
  }
\end{figure}
The simple robot experiment (see Figure~\ref{\LabelObjectHere{Simple Robot Run and Trees}}) scales the blackboard while the tree is constant\@.
A robot on a $n$ by $n$ board tries to reach a goal\@.
Once reached, a new goal is generated\@.
% This process repeats 3 times\@.
We verify that eventually 3 goals are reached\@.
The experiment scales by increasing $n$ from $2$ to $30$ in increments of 4\@.
We compare to MoVe4BT for this experiment\@.
MoVe4BT has no concept of an environment so the environment has to become part of the $\BehaviorTree$\@.
Furthermore, MoVe4BT cannot assign values to variables nondeterministically; instead MoVe4BT nodes can nondeterministically choose one of $\Success$, $\Running$, $\Failure$, and a single accompanying `program' with updates\@.
Thus to achieve the effect of nondeterminism, we had to use a series of nodes\@.
Each node could nondeterministically choose to change the value or not\@.
Thus the tree is more complex and also grows in size with the blackboard\@.
Finally, we had to introduce an extra node (called $MoVe4BTCheck$) so that MoVe4BT could verify the condition at all\@.
Refer to Figure~\ref{\LabelObjectHere{Simple Robot Run and Trees}} for a visual comparison of the trees\@.
See Figure~\ref{\LabelObjectHere{Simple Robot Timing Results}} for timing results\@.
\begin{figure}%[b!]
  %\vspace{-15pt}
  \begin{minipage}{.47\linewidth}
    \includegraphics[width=\linewidth,keepaspectratio]{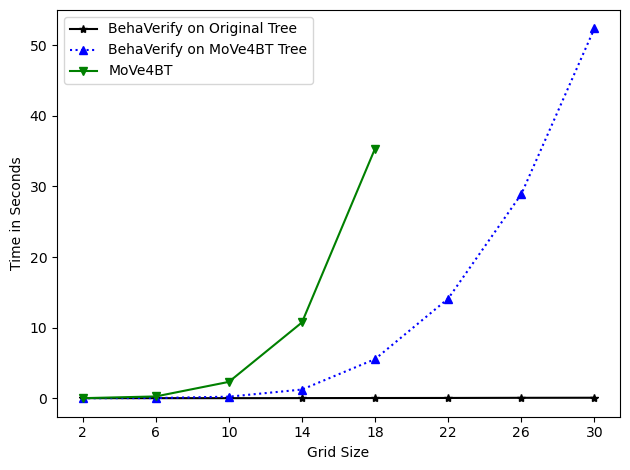}
  \end{minipage}
  \begin{minipage}{.52\linewidth}
    \caption{
      Time in seconds to verify that finally, there are no remaining goals: $\LTLFinally(RG = 0)$\@.
      Our tool ran on both the original tree and the changed tree\@.
      %The original tree verified the condition directly while the changed tree required additional nodes to be inserted\@.
      Starting at a 22 by 22 grid, MoVe4BT ran for over a minute and then produced a blank screen; we interpreted this as a timeout\@.
      At 18 by 18, BehaVerify reported about $2^{19}$ reachable states\@. 
      At 30 by 30, BehaVerify reported about $2^{22}$ reachable states\@.
    }\label{\LabelObjectHere{Simple Robot Timing Results}}
  \end{minipage}
  %\vspace{-20pt}
\end{figure}
\NewParagraph{Moving Target Experiment}
For this experiment, a drone is searching for a mobile target on a grid\@.
The drone has limited vision, which is further obscured by trees (the trees do not prevent movement)\@.
\begin{figure}%[t!]
%\vspace{-2em}
  \begin{minipage}{.3\linewidth}
    \includegraphics[width=\linewidth,keepaspectratio]{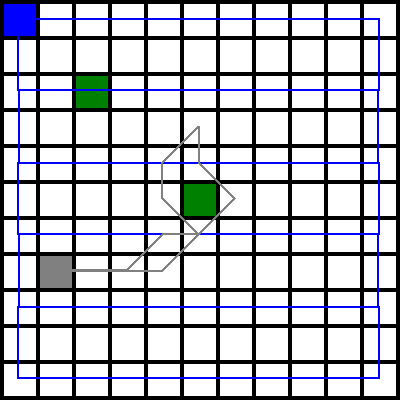}
  \end{minipage}
  \begin{minipage}{.68\linewidth}
    \caption{
      The target (gray) is able to avoid the drone (blue) by circling a tree (green) that obstructs vision but not movement\@.
      The target's initial position and movement are nondeterministic; the drone's are not\@.
      Note that this was a counterexample generated by nuXmv from the generated model\@.
    }\label{\LabelObjectHere{ISR Evading Trace}}
  \end{minipage}
  %\vspace{-15pt}
\end{figure}
We conduct two experiments for this setup: one where the target can move every 5 turns and one where the target can move every 10 turns\@.
When the target can move every 10 turns, we verify that eventually the drone finds the target\@.
When the target can move every 5 turns, we generate a counterexample (see Figure~\ref{\LabelObjectHere{ISR Evading Trace}})\@.
In both cases, the specification being verified is that we eventually `see' the target\@.
Given the complexity of encoding the simple robot environment for MoVe4BT and the fact that this experiment is more complex and nondeterministic, we did not attempt to recreate it in MoVe4BT\@.
\NewParagraph{Reasoning About Results}
We believe fastforwarding (see Section~\ref{\LabelSection{\CurrentChapter}{Fastforwarding Execution}}) is the main reason BehaVerify outperforms MoVe4BT\@.
Based on the traces that MoVe4BT produces, we believe that MoVe4BT jumps from leaf to leaf during execution\@.
This provides a performance boost over stepping through every single node in the tree, but it is not enough\@.
Consider the bigger fish experiment\@.
With 20000 nodes, there are 19997 leaf nodes\@.
During the first tick, 2 leaf nodes will be active\@.
During the second tick, there will be 3\@.
Finally, there will be 19996 active leafs\@.
Thus, MoVe4BT would have to step through $2 + 3 + \cdots + 19996 = 199,930,005$ leaf nodes resulting in a very long trace\@.
By comparison, the BehaVerify trace would have less than 20000 states\@.
Next, we consider the simple robot experiment, where changing the tree had a huge impact\@.
This is because MoVe4BT does not support nondeterministic variable assignments; as such we had to create a series of nondeterministic nodes\@.
Thus instead of having one nondeterministic update for $x$ and one for $y$, we had $n-1$ for each\@.
This caused the number of total states to jump from $2^{23}$ to $2^{80}$ while the number of reachable states was relatively unchanged\@.
The changed tree is close to a worst case scenario for BehaVerify; many variable updates, all of them nondeterministic\@.
Even in this worst case scenario, BehaVerify significantly outperformed MoVe4BT\@.

%                                                                                                          4.0 pages
\NewSection{Conclusions and Future Work}
We introduced and formally defined $\StatefulBehaviorTree$s and demonstrated they are Turing Complete under certain assumptions\@.
We presented a DSL for specifying $\StatefulBehaviorTree$s implemented in BehaVerify\@.
Our experiments demonstrate BehaVerify can complete a verification task on a tree with 20000 nodes in the time that MoVe4BT, a different verification tool, verifies a property on a tree with 200 nodes, demonstrating two orders of magnitude scalability improvement\@.
Potential future work includes improving our encoding of array variables, developing a graphical user interface for the creation of $\StatefulBehaviorTree$s, visualization of counterexamples provided by nuXmv, general performance improvements, and expanding support for unbounded variable types in nuXmv for bounded model checking (BMC).

%                                                                                                          0.5 pages
\section*{Acknowledgements}
The material presented in this paper is based upon work supported by the National Science Foundation (NSF) through grant numbers 2220426 and 2220401, the Defense Advanced Research Projects Agency (DARPA) under contract number FA8750-23-C-0518, and the Air Force Office of Scientific Research (AFOSR) under contract numbers FA9550-22-1-0019 and FA9550-23-1-0135. This paper was also supported in part by a fellowship award under contract FA9550-21-F-0003 through the National Defense Science and Engineering Graduate (NDSEG) Fellowship Program, sponsored by the Air Force Research Laboratory (AFRL), the Office of Naval Research (ONR), and the Army Research Office (ARO). Any opinions, findings, and conclusions or recommendations expressed in this paper are those of the authors and do not necessarily reflect the views of AFOSR, DARPA, or NSF.

%                                                                                                          0.0 pages     <- this doesn't need to be included yet
%                                                                                                          17.5 pages without references
% references
%                                                                                                          2.0 pages
%                                                                                                          19.5 pages with references

\bibliographystyle{eptcs}
\bibliography{./Utility/Bibliography}

\end{document}